# XuanJian-DINOSAR: A SAR Foundation Model Based on Global and Local Semantic Consistency

Chunxiao Wu, Huixin Huang, Jialei Pan, Qiang Liang, Tao Wu, Cheng Wu,

Long Huang, Yuanji Li, Qiang Cheng, Aifang Liu, Linghao Xia, Rui Zhang*

**Abstract—Self-supervised pre-training on synthetic aperture radar (SAR) imagery provides transferable representations for classification and detection, yet representation stability remains constrained by the imaging-domain gap between optical and SAR data and by speckle-induced perturbations. Most existing SAR foundation models follow the masked image modeling (MIM) paradigm and learn representations through pixel reconstruction or structural feature prediction. Pixel reconstruction, however, is sensitive to speckle and intensity fluctuations, whereas structural feature prediction depends on predefined feature operators. Moreover, MIM-based methods primarily emphasize local semantics, leaving the joint modeling of cross-view semantic relations and perturbation stability at the same spatial location insufficiently explored in the SAR domain. To address this issue, we propose XuanJian-DINOSAR, a SAR foundation model based on global and local semantic consistency. We first curate more than 360K+ publicly available SAR target chips, detection images, and scene images to construct an unlabeled pre-training collection covering diverse targets, scenes, and sensors. Starting from an optically pre-trained DINOv2 model, we perform self-supervised domain adaptation on SAR imagery. The resulting framework combines DINO global semantic self-distillation, iBOT local masked modeling, and KoLeo feature-distribution regularization to learn general-purpose representations with both global semantic discrimination and local structural perception. We further introduce a local semantic consistency branch under perturbations to account for the intensity statistics and coherent imaging characteristics of SAR data. The branch generates paired teacher weak views and student strong views through shared geometric transformations, applies intensity transformations and Gamma multiplicative speckle perturbations to the student views, and imposes a cosine consistency constraint on patch tokens at corresponding spatial locations. It therefore encourages local representations to remain consistent under radiometric perturbations. This objective complements the cross-region and cross-scale semantic consistency learned by DINOv2 through multi-crop training, allowing the model to learn global and local semantic consistency jointly. Extensive experiments on multiple downstream classification and detection datasets show that XuanJian-DINOSAR achieves state-of-the-art performance on most evaluation metrics, supporting the generality and transferability of the learned foundation-model representations.**

This work was supported in part by the National Natural Science Foundation of China (NSFC) under Grant No. 62427801 (National Major Scientific Research Instrument Development Project) and in part by the Joint Fund for Enterprise Innovation and Development of the National Natural Science Foundation of China (NSFC) under Grant No. U25B6001.

The authors are with the Nanjing Institute of Electronic Technology and National Key Laboratory of Radar Detection and Sensing, Nanjing 210039, China.

## I. INTRODUCTION

Synthetic aperture radar (SAR) acquires observations by actively transmitting microwaves and receiving backscattered echoes from the scene. Its day-and-night, all-weather imaging capability makes it an important sensing modality for ocean monitoring, disaster assessment, and surface-target interpretation. Unlike optical imaging, which relies on reflected radiance, SAR measurements result from the coherent superposition of scattering returns within each resolution cell; the observed intensity is therefore governed jointly by target scattering properties and speckle [1]. Variations in sensor configuration, imaging geometry, and acquisition conditions further alter the scattering response of a target and its relationship with the background. As a result, instances of the same class may exhibit large appearance variations, while discriminative structures between classes may be obscured by clutter and intensity fluctuations [2]. SAR representation learning must consequently capture globally discriminative information, preserve sensitivity to local scattering structures, and reduce sensitivity to nonsemantic radiometric perturbations. These requirements become particularly important when a representation is expected to transfer across multiple data sources and downstream tasks.

Deep learning has shifted SAR target classification and detection from handcrafted feature design toward data-driven representation learning. Chen et al. [3] used a fully convolutional network to learn hierarchical SAR target features, reducing dependence on handcrafted descriptors. An et al. [4] proposed DRBox-v2, which combines rotated bounding boxes with multi-level features to improve localization across target scales. Xia et al. [5] introduced CRTransSar, integrating convolution and Transformers through contextual joint representation and cross-resolution feature interaction for ship detection. These studies show that local structure extraction, global context modeling, and multi-scale feature fusion are central to SAR interpretation. Task-specific models, however, are generally optimized for fixed category sets, datasets, and supervision objectives, and their performance depends on the availability of accurate annotations and sufficient coverage of the training distribution. Changes in target category, sensor, or task often require additional labels and parameter adaptation [2], [6]. As the amount of accessible SAR data continues to grow, learning reusable representations for classification and detection from unlabeled imagery has become a key direction for moving beyond task-specific learning.

Large-scale self-supervised pre-training offers a practical route toward this goal. Masked autoencoders (MAEs) learn contextual relations by recovering occluded image content [7], whereas DINO aligns predictions across views through teacher–student self-distillation and organizes visual features semantically without class labels [8]. DINOv2 further integrates image-level self-distillation, patch-level masked prediction, and feature-distribution regularization, providing a unified representation basis for image-level discrimination and local structure perception [9]. In remote sensing, SatMAE extends masked pre-training to temporal and multispectral imagery [10], RingMo learns generic remote-sensing features through masked image modeling [11], and CROMA combines radar–optical contrastive learning with masked reconstruction for both unimodal and multimodal representations [12]. Despite this transfer potential, visual priors learned from natural or optical remote-sensing images do not directly remove the domain gap induced by SAR imaging. Through feature analysis on the SAMPLE task, Inkawhich [13] found that DINOv2 without SAR adaptation tends to separate simulated from measured samples, whereas continued self-supervised training on SAR imagery improves class separability and cross-domain correspondence. Their comparison of several downstream adaptation strategies further indicates that SAR-domain self-supervision is an important step when reusing existing visual foundation models and cannot be replaced merely by increasing model scale.

To accommodate SAR imaging characteristics, recent foundation-model studies have developed several domain-specific pre-training schemes, most of which are based on masked image modeling (MIM). SARATR-X [6] adopts two-stage self-supervised pre-training and multi-scale gradient-feature supervision to extend general representation learning to SAR target classification and detection. SUMMIT [14] incorporates self-supervised denoising and spatial scattering-feature enhancement into masked modeling and coordinates multiple supervision signals through an auxiliary-task mechanism. SAMBA [15] combines a bidirectional Mamba encoder, hierarchical scattering-guided masking, and lightweight spatial feature interaction to balance computational efficiency with the modeling of SAR scattering structures. SARMAE [16] constructs a perturbation-reconstruction task through speckle-aware augmentation and uses paired optical imagery with a frozen optical encoder to provide patch-level semantic anchors. SARATR-X-v2 [17] further considers both physical stability and scale compatibility in its reconstruction target, using learnable weights to fuse fixed structural responses across multiple receptive fields into a unified supervision signal. These methods improve SAR pre-training through complementary choices of supervision target, auxiliary task, masking strategy, and semantic source, showing that both domain priors and objective design affect representation quality. AFRL-DINOv2 [13], meanwhile, demonstrates the feasibility of continued teacher–student self-distillation in the SAR domain and provides an alternative route based on learned semantic targets.

Although these methods have advanced SAR pre-training, their supervision mechanisms address different conditions. Pixel reconstruction directly fits measurements affected by speckle and intensity fluctuations, so its objective may entangle target structure with unstable radiometric variation [16], [17]. Structural-feature supervision reduces this sensitivity, but its representational bias remains constrained by the selected operators; even with learnable cross-scale fusion, the structural extraction procedure must be specified in advance [6], [17]. Auxiliary denoising and paired cross-modal alignment broaden the supervision sources, but the former is centered on recovering perturbed images, whereas the semantic-anchor branch of the latter requires spatially corresponding optical observations [14], [16]. These mechanisms can improve representation learning, yet they do not directly constrain learned SAR features at the same spatial location under different perturbations. In DINOv2, multi-crop self-distillation establishes semantic relations across regions and scales, while iBOT provides local predictive supervision at masked positions [9]. Neither objective explicitly enforces consistency between weakly and strongly perturbed patches under shared geometry. We therefore ask how SAR-domain adaptation can preserve cross-region and cross-scale semantic learning while explicitly stabilizing local representations of the same SAR region under intensity variation and simulated speckle, and how these two objectives can be made complementary.

We address this question with XuanJian-DINOSAR, a SAR foundation model based on global and local semantic consistency. More than 360,000 public SAR target chips, detection images, and scene images are curated into an unlabeled pre-training collection spanning diverse targets, scenes, and sensors. An optically pre-trained DINOv2 ViT-B/14 initializes self-supervised domain adaptation on this collection. The framework retains DINO global semantic self-distillation, iBOT local masked modeling, and KoLeo feature-distribution regularization [9] to learn general-purpose representations with global semantic discrimination and local structural perception. We additionally introduce a local semantic consistency branch under perturbations. The branch reuses the geometric transformations of the global views to construct a teacher weak view and a student strong view of the same SAR region. Intensity transformations and Gamma multiplicative speckle perturbations are applied to the student view, while an exponential-moving-average teacher supplies local feature targets for a cosine consistency constraint between corresponding patch tokens. The branch requires only SAR imagery during domain adaptation, and its supervision is provided by learned teacher features from the same modality. By focusing on perturbation stability at corresponding spatial locations, it complements the cross-region and cross-scale semantic consistency learned by DINOv2 through multi-crop training and jointly exploits cross-view semantic relations and local spatial correspondence.

The main contributions are summarized as follows:

1) We develop XuanJian-DINOSAR, a self-supervised domain-adaptation framework for SAR imagery. By jointly using the image-level and patch-level learning mechanisms of DINOv2 on multi-source unlabeled SAR data, the framework learns a reusable backbone that combines global semantic discrimination with local structural perception for classification and detection.

2) We introduce a shared-geometry weak–strong local consistency branch. Feature constraints between spatially corresponding teacher and student patches explicitly promote local semantic consistency under intensity variation and multiplicative speckle perturbations, while being optimized jointly with the original cross-region and cross-scale multi-crop objectives of DINOv2.

3) We conduct extensive evaluations on multiple downstream classification and detection datasets, including XuanJian-RSAR [28], our oriented fine-grained target-detection dataset. XuanJian-DINOSAR achieves state-of-the-art results on most reported metrics, supporting the generality and transferability of the proposed foundation model.

## II. RELATED WORK

### *A. Task-Specific SAR Models*

SAR target classification typically takes target chips as input and learns discriminative features over a predefined category set. Early deep-learning methods mainly replaced handcrafted feature extractors with convolutional networks. The fully convolutional network of Chen et al. [3], for example, removes fully connected layers and retains hierarchical feature extraction with fewer parameters, thereby reducing overfitting under limited annotations. Such models progressively aggregate local scattering responses into target-level representations and laid the foundation for end-to-end SAR recognition. As research moves toward more complex imaging conditions, classification representations must also accommodate distribution shifts induced by changes in aspect, depression angle, background, and sensor [2]. A decision boundary learned on a fixed category set and data split does not guarantee transfer to new classes or acquisition conditions. Preserving reusable target-structure information therefore remains a central problem in SAR recognition [6].

Compared with chip-based classification, SAR object detection requires both category discrimination and spatial localization and is therefore more dependent on small-target detail, multi-scale context, and orientation information. DRBox-v2 [4] represents objects with rotated bounding boxes and combines multi-level anchors with a feature pyramid, allowing features at different levels to contribute to localization across target scales and orientations. CRTransSar [5] further combines Transformer-based context modeling with convolutional local feature extraction through a contextual joint-representation backbone and cross-resolution attention enhancement for multi-scale ship detection. These methods improve feature extraction and fusion within task-specific detectors, indicating that effective SAR detection representations must retain both local structure and scene context; image-level category information alone is insufficient for spatial localization.

The feature learning of task-specific models is closely tied to class labels, object boxes, and the training distribution. Architectural improvements can strengthen representations for a particular task, but cross-task reuse still depends on the knowledge encoded during pre-training and on the downstream adaptation strategy [2], [6]. Classification emphasizes target-level semantic discrimination, whereas detection additionally requires spatial correspondence. The two tasks thus share a need for generic features but differ in representational granularity. Our pre-training framework considers both requirements by combining image-level semantic learning with local representation constraints, leaving task-specific adaptation to the downstream stage.

### *B. Vision and Remote-Sensing Foundation Models*

Vision foundation models decouple feature learning from fixed-category supervision through large-scale self-supervised pre-training and provide shared representations for multiple tasks. Masked image modeling predicts occluded content from visible regions, encouraging the model to capture relations between local information and context. MAE [7] uses an asymmetric encoder–decoder: the encoder processes only visible patches, and a lightweight decoder reconstructs the masked pixels, reducing pre-training cost while learning transferable features. The supervision target in this paradigm can be defined in pixel space or generated by a feature extractor. What the model predicts is therefore as consequential as the masking strategy itself. For SAR imagery with multiplicative speckle, raw pixels contain both target structure and radiometric responses that vary with acquisition conditions, so target design must account for domain-specific statistics [1], [17].

Teacher–student self-distillation constructs learning signals by matching predictions across views. DINO [8] uses a momentum-updated teacher and multi-crop training: the student processes both global and local views and learns to match teacher predictions from the global views, thereby linking local and global semantics. iBOT [18] further treats the teacher as an online tokenizer and supervises student predictions at masked patch positions in addition to cross-view image-level distillation, allowing local targets to evolve with the representation. DINOv2 [9] combines the image-level DINO objective with the patch-level iBOT objective and feature-distribution regularization, yielding a pre-training framework for both image-level discrimination and dense representation. These studies show that cross-view semantic learning and local prediction can jointly shape general-purpose features. Building on the positional supervision already provided by masked patch prediction, our work further targets feature stability under SAR-specific radiometric perturbations.

Remote-sensing foundation models combine these pre-training mechanisms with temporal, spectral, and multimodal properties of Earth-observation imagery. SatMAE [10] introduces temporal encoding and independent masking across time and uses band grouping with spectral positional embeddings to exploit observation dimensions absent from natural images. RingMo [11] learns generic remote-sensing features through domain-scale masked modeling. CROMA [12] separately encodes spatiotemporally aligned SAR and optical images, combines cross-modal contrastive learning with masked reconstruction, and obtains multimodal representations through a fusion encoder. Remote-sensing pre-training thus requires both the transfer of visual learning mechanisms and adaptation to modality-specific data and acquisition conditions. Our approach initializes from an optically pre-trained model, continues representation learning in the SAR domain, and constructs local consistency supervision from paired augmented views of the same SAR sample.

### *C. SAR Foundation Models*

A major direction in SAR foundation models is to incorporate imaging priors into self-supervised objectives so that the learned representations better match SAR data [2]. Current approaches are largely organized around masked image modeling (MIM). For structural-feature supervision, SAR-JEPA [19] predicts multi-scale ratio-gradient features through local masked modeling, using domain priors to reduce the influence of multiplicative speckle on supervision and evaluating transfer across vehicle, ship, and aircraft classification. SARATR-X [6] follows the multi-scale gradient-supervision principle and combines a HiViT backbone with two-stage self-supervised pre-training from natural images to SAR imagery, extending the resulting model to target classification and detection. SARATR-X-v2 [17] formulates target design as the joint problem of physical stability and scale compatibility. It uses fixed structural extractors spanning six receptive fields and learnable cross-scale weights to construct a unified reconstruction target. This line of work marks a progression from predicting raw observations toward predicting stable structure. Its domain knowledge enters primarily through structural operators and their composition, whereas the additional local target in our method is supplied dynamically by a teacher network trained in the SAR domain.

A second family of methods enriches masked pre-training through auxiliary tasks and external semantic information. SUMMIT [14] incorporates self-supervised denoising and enhancement of edge and scattering-point features into MIM and coordinates these signals with an auxiliary-task mechanism, allowing the model to use both image recovery and spatial scattering priors. SARMAE [16] injects speckle perturbations into SAR inputs and reconstructs the original observations as a denoising masked objective. For samples with paired optical observations, it also uses a frozen DINOv3 optical encoder as a semantic anchor and imposes cosine-distance constraints between corresponding visible patches. This approach combines speckle augmentation with local semantic alignment, but its semantic-anchor branch depends on cross-modal registration. Our method also targets local representation consistency, yet the correspondence is generated by shared geometric transformations of the same SAR image. The teacher is updated by exponential moving average during SAR-domain training, so the local supervision directly addresses within-modality intensity changes and simulated speckle perturbations.

Beyond target design, backbone architecture and mask allocation provide additional routes for introducing SAR priors. SAMBA [15] processes image tokens with a bidirectional Mamba encoder and constructs hierarchical scattering-guided masks through regional-density adjustment, multi-scale block-structure constraints, and scattering-point contrast modulation, while SpatialMix provides lightweight spatial feature interaction. The resulting mask allocation follows the spatial distribution of target scattering information while controlling the cost of sequence modeling. Such methods primarily reorganize visible information and its encoding process. We retain the DINOv2 backbone and base objectives and instead focus on the perturbation design of paired views and explicit constraints between their corresponding local representations. The emphasis therefore differs from scattering-guided masking and backbone redesign.

For self-distillation in the SAR domain, AFRL-DINOv2 [13] starts from an optically pre-trained DINOv2 model and continues self-supervised training on a curated SAR collection. In addition to analyzing the effect of pre-training data composition, the study compares nearest-neighbor classification, linear and multilayer probing, LoRA, and full fine-tuning, demonstrating the transfer value of learned teacher targets in the SAR domain. AFRL-DINOv2 retains the original DINOv2 objectives with domain-specific augmentation. By contrast, our framework explicitly constructs weak–strong views that share the complete geometric transformation and imposes a cosine consistency constraint on patches at the same spatial location under intensity and Gamma multiplicative-speckle perturbations. We evaluate the resulting representation on both classification and detection.

Stable structural targets, scattering-aware pre-training tasks, and learned semantic supervision thus provide different constraints for general-purpose SAR representation learning. We combine the cross-region and cross-scale semantic relations established by DINOv2 multi-crop training with local semantic consistency at the same spatial position under SAR perturbations. The former connects semantic information across cropped views, whereas the latter constrains changes in features from geometrically corresponding regions. The new branch is trained jointly with iBOT masked prediction so that cross-view semantic learning and local perturbation stability complement each other within a unified teacher–student framework.

## III. METHOD

### *A. Overall Framework*

XuanJian-DINOSAR performs self-supervised domain adaptation on unlabeled SAR imagery to learn a transferable generic SAR encoder from optically pre-trained representations. The pre-training data are curated from public SAR datasets and include target chips, detection images, and scene images; their sources and composition are detailed in Section IV. Data curation follows the processing pipeline of SUMMIT [14], including image cleaning, duplicate removal, and input-size normalization. ORB local-feature matching is used to screen for duplicate and redundant images. The data partition is determined jointly with the downstream evaluation protocol: pre-training includes only samples permitted for training, while test images and their duplicate or near-duplicate derivatives are excluded. For chips traceable to the same source image, overlap across splits is checked at the source-image level to prevent adjacent or repeated regions from being assigned separately to the pre-training and test sets. This test-isolation principle applies independently of label usage and is not relaxed because pre-training itself is unlabeled.

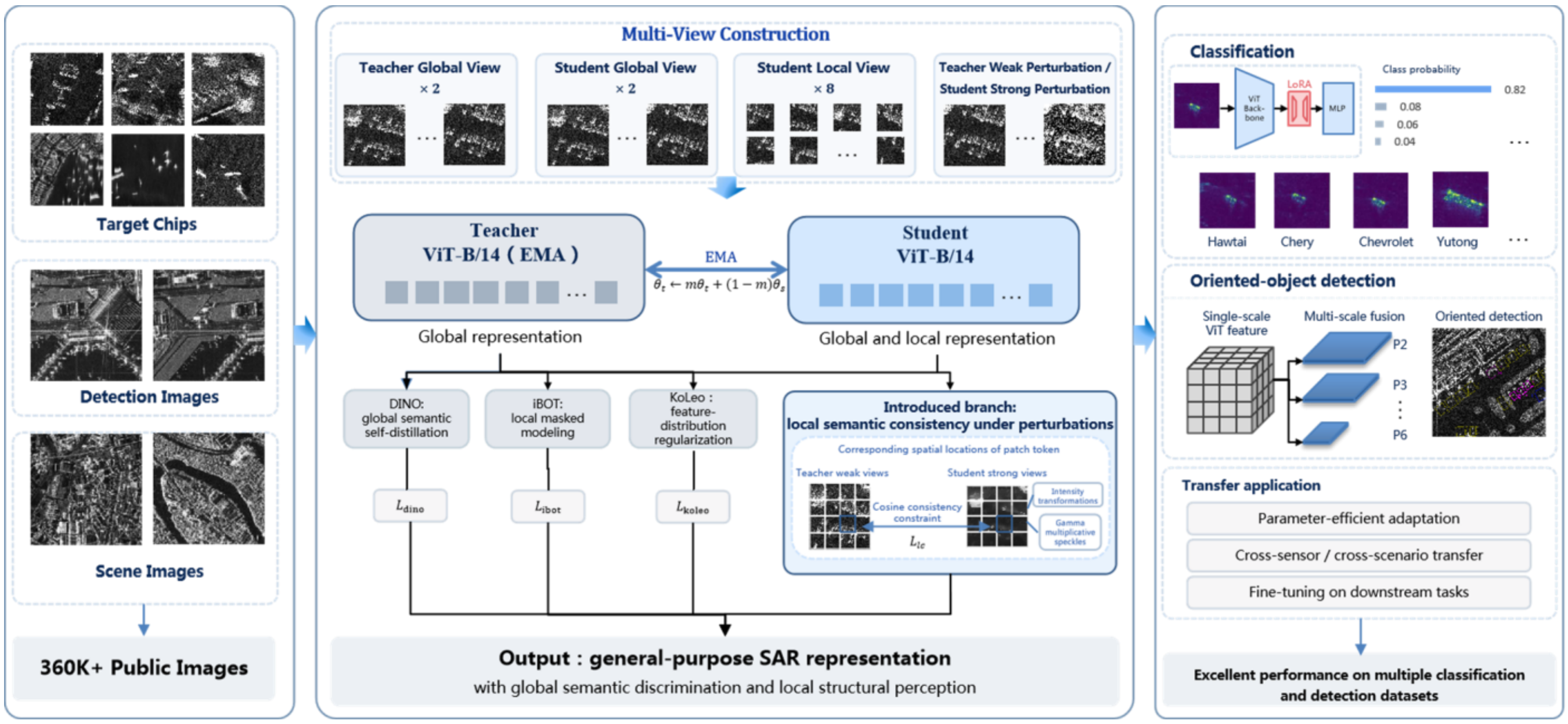

Fig. 1. Overall architecture of XuanJian-DINOSAR. the framework contains a conventional multi-crop self-distillation pathway and a shared-geometry weak–strong consistency pathway. The former preserves the DINOv2 mechanisms for global semantic prediction, local masked prediction, and feature-distribution regularization, establishing semantic relations across regions and scales. The latter reuses the global crop and changes the image-intensity statistics under an identical geometric layout, explicitly constraining teacher and student patch features at corresponding spatial locations. After pre-training, classification heads and detection networks are adapted separately for downstream tasks.

The backbone is DINOv2 ViT-B/14 with register tokens. Structurally identical student and teacher encoders are initialized from optically pre-trained weights [9]. Each image is divided into nonoverlapping 14 × 14 patches and linearly embedded before being processed by the Transformer together with one CLS token and four register tokens. For a 224 × 224 global view, the spatial grid contains 16 × 16 patch tokens. The CLS token aggregates view-level context and is used by DINO cross-view prediction and KoLeo regularization. Patch tokens retain spatial representations aligned with the input grid and are used by iBOT masked prediction and the proposed local-consistency constraint. Register tokens act as learnable auxiliary tokens during feature interaction, do not correspond to physical image patches, and are excluded from position-wise losses.

As illustrated in Fig. 1, the framework contains a conventional multi-crop self-distillation pathway and a shared-geometry weak–strong consistency pathway. The former preserves the DINOv2 mechanisms for global semantic prediction, local masked prediction, and feature-distribution regularization, establishing semantic relations across regions and scales. The latter reuses the global crop and changes the image-intensity statistics under an identical geometric layout, explicitly constraining teacher and student patch features at corresponding spatial locations. Both pathways share the student and teacher encoders. Their gradients update the student jointly, whereas the teacher provides stop-gradient targets. DINO and iBOT operate on prediction distributions produced by their projection heads, while KoLeo and the local-consistency loss act on backbone representations. After pre-training, the exponential-moving-average (EMA) teacher encoder is retained as the foundation model; classification heads and detection networks are adapted separately for downstream tasks.

### *B. Multi-View SAR Data Augmentation*

The conventional multi-crop pathway constructs two global views and eight local views from each SAR image. Global crops cover larger regions and preserve the contextual relationship between targets and background. Local crops retain scattering structures within smaller regions, requiring the student to predict the global semantic distribution supplied by the teacher from limited observations. Each view undergoes random scale cropping and resizing, horizontal and vertical flipping, and rotations by multiples of 90°, together with Gaussian blur and intensity augmentation. Crop scales provide cross-region and cross-scale self-supervision, while flips and rotations increase directional variation in the image plane. These transformations should not be interpreted as physical reconstructions of the true incidence angle, azimuth angle, or sensor trajectory.

The additional consistency pathway constructs a weak–strong pair from each of the two global crop regions. Let a batch contain $B$ images and $G = 2$ shared global views, with $T_g$ denoting the complete geometric transformation for the gth global view. For each sample, the two branches share all crop, resize, flip, and rotation parameters and use the same spatial patch ordering. Let $u$ denote the nonnegative image after the geometric transformation. The teacher weak view applies only basic numerical processing and input normalization. The student strong view sequentially applies ClipAndScale, a power-law transformation, and Gamma multiplicative perturbation, followed by the same input normalization. All intensity augmentations are performed in the nonnegative domain before standardization. This ordering avoids exponentiating standardized negative values or misinterpreting a zero-mean standardized quantity as radar intensity. The weak and strong branches use the same normalization rule after their respective intensity processing. Because they share the complete geometric transformation and the strong branch changes only the intensity statistics, patch-wise spatial correspondence is preserved.

ClipAndScale linearly rescales values between selected lower and upper thresholds and clips extreme responses outside this interval:

$$C_{a,c}(u) = \frac{\min\{\max(u,a),c\} - a}{c-a}, 0 \le a < c. \quad (1)$$

where $u$ is the nonnegative value of an individual pixel, and $a$ and $c$ are the lower and upper clipping thresholds, respectively. The output lies in [0, 1]. Sampling the thresholds from a predefined augmentation range changes the visible dynamic range, reduces the dominance of a small number of very strong scatterers over the numerical scale, and generates observations under different contrast conditions. This operation changes the numerical mapping of an image and should not be interpreted as absolute radiometric calibration across sensors. The sampling range and selection rule for the thresholds are treated as augmentation hyperparameters.

The power-law transformation further adjusts the relative contrast between weak- and strong-scattering regions after clipping and rescaling:

$$P_\gamma(u) = u^\gamma, u \in [0,1], \gamma > 0. \quad (2)$$

where $\gamma$ is a randomly sampled exponent. Values of $\gamma$ below 1 raise low-intensity responses, whereas values above 1 suppress them, thereby changing the visible contrast among local scattering responses. This monotonic transformation preserves intensity ordering while exposing the model to nonlinear intensity mappings so that representation learning is not tied to a single display or preprocessing condition. The sampling range of $\gamma$ remains configurable.

Gamma multiplicative speckle perturbation modulates the image pixel-wise with a unit-mean random field:

$$[S_L(u)]_{ij} = u_{ij}\xi_{ij}, \xi_{ij} \sim \text{Gamma}(N, 1/N), N > 0. \quad (3)$$

$$E[\xi_{ij}] = 1, \text{Var}(\xi_{ij}) = \frac{1}{N}. \quad (4)$$

where $i$ and $j$ index pixel coordinates and the product of $u_{ij}$ and $\xi_{ij}$ denotes pixel-wise multiplicative modulation. The Gamma distribution uses the shape-scale parameterization, and $N$ controls the perturbation strength. Smaller values of $N$ produce larger multiplicative fluctuations, whereas larger values produce weaker random modulation. Under an ideal multi-look intensity-speckle model, $N$ is related to the equivalent number of looks [1]; in this work, $N$ is first treated as a statistical parameter controlling augmentation strength. For public SAR imagery that has been quantized, compressed, or nonlinearly mapped, the perturbation approximates multiplicative fluctuations but does not reproduce the complete coherent-imaging process, spatial speckle correlation, or sensor postprocessing pipeline. The Gamma perturbation is applied after the power-law transformation; its output may exceed 1 and is subsequently normalized together with the other inputs.

### *C. Teacher–Student Self-Distillation*

The base training pathway follows the DINOv2 teacher–student self-distillation mechanism [9]. The student is optimized through backpropagation, whereas the teacher receives no gradients. Its backbone and corresponding projection heads are updated through an exponential moving average of the student parameters. Let $\theta_s$ and $\theta_t$ denote the student and teacher parameters at iteration k. After the student update, the teacher parameters are computed as:

$$\theta_t^{(k+1)} = m_k\theta_t^{(k)} + (1-m_k)\theta_s^{(k+1)}. \quad (5)$$

where $m_k$ is the EMA momentum, which gradually increases toward 1 during training and yields smoother teacher targets in later stages. Stop-gradient is applied to both teacher prediction distributions and local features; consequently, the teacher is updated only by EMA and receives no backpropagated gradients from the DINO, iBOT, or LC loss. The two training pathways use the same EMA teacher, so the additional consistency constraint does not require a separate target network.

For cross-view prediction, the teacher receives two global views, while the student receives all global and local views. The CLS representation of each view is mapped by the DINO projection head to a prediction distribution. Teacher distributions from the global views guide student predictions on the remaining views; the pairing between a teacher global view and the identical student global view is excluded from the DINO loss. This process links student representations from local regions to teacher semantics derived from a broader context and supports cross-region and cross-scale information exchange. DINO learns through self-distilled distribution matching and does not require explicit negative samples. Its anti-collapse mechanisms include centering and temperature control of teacher targets together with jointly optimized feature-distribution regularization.

For local masked learning, a subset of patch embeddings in each student global view is replaced by mask tokens while their sequence positions are retained. The teacher processes the corresponding unmasked view. iBOT uses online target distributions from the teacher to supervise student predictions at masked positions, requiring the student to recover local representations from visible context [18]. The recovery target is the teacher patch-prediction distribution rather than raw SAR pixels. DINO and iBOT therefore provide view-level and local position-level supervision, respectively: DINO acts on CLS tokens, whereas iBOT acts on patch tokens at masked positions.

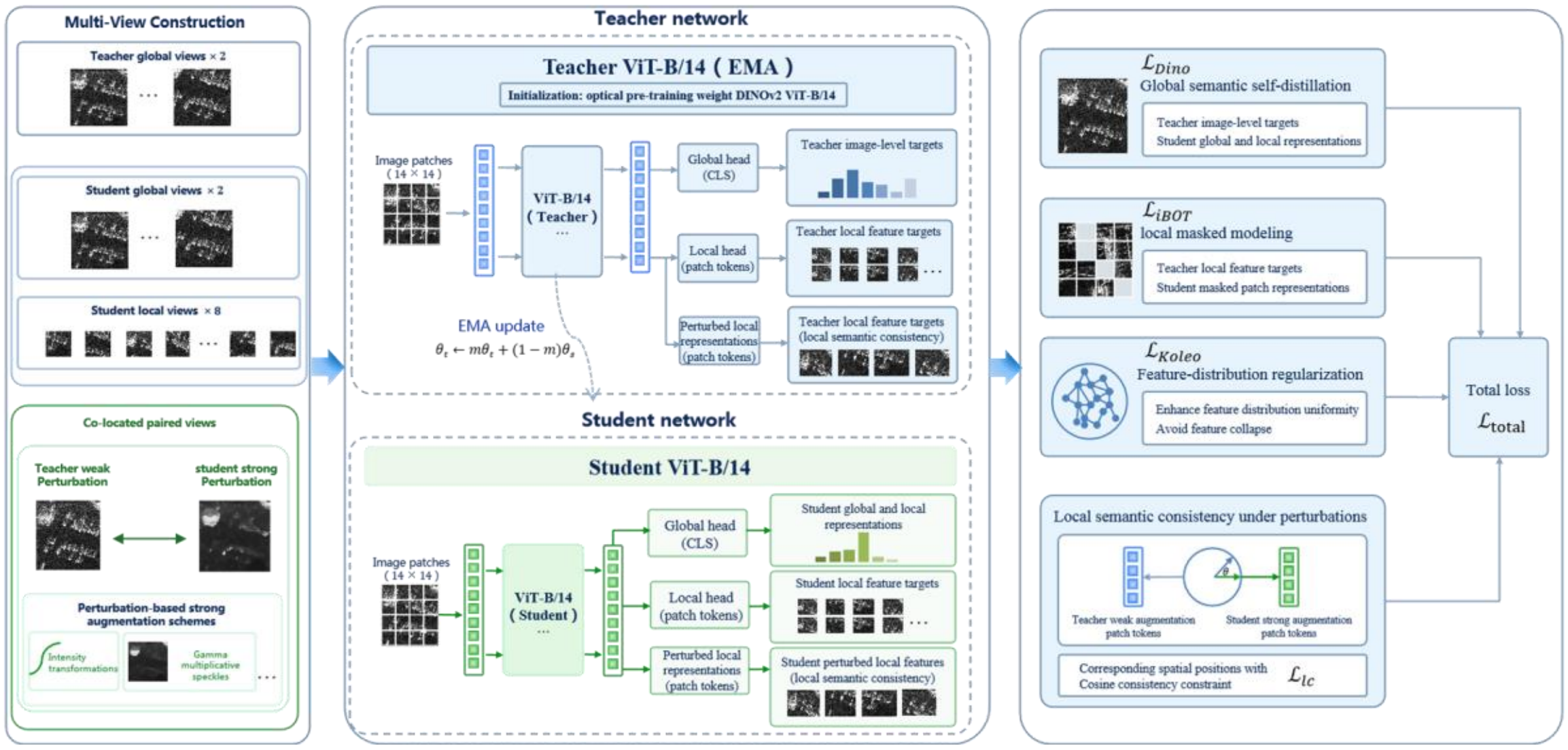

Fig. 2. SAR self-supervised architecture with global semantic discrimination and local perturbation-consistent representations. The cosine loss in the diagram corresponds to the local-consistency loss defined in the text.

KoLeo regularization acts directly on global CLS features from the student backbone. After feature-wise normalization, it encourages representations of different images to remain appropriately dispersed within a batch [9]. The two global-view sets are processed separately so that the two views of the same sample are not treated as neighbors to be repelled. KoLeo neither operates on register tokens nor uses teacher prediction distributions. These mechanisms are inherited from DINOv2. The proposed branch shares their encoders and adds a cross-perturbation constraint at corresponding spatial locations to the same joint optimization, as shown in Fig. 2.

### *D. Local Consistency Learning under Speckle and Intensity Perturbations*

Multi-crop self-distillation requires views with different spatial coverage to establish semantic relations, but its view-level objective does not directly specify how the same local region should be represented under intensity changes and additional multiplicative fluctuations. iBOT complements this objective by predicting teacher distributions at masked positions, with supervision concentrated on the masked locations. Motivated by this distinction, we introduce a local-consistency branch that establishes an explicit position-wise feature constraint between weak and strong views sharing the same geometry. The model therefore receives both cross-crop semantic supervision and same-location perturbation-consistency supervision.

For the $g$th shared global region of sample $b$, the teacher receives the weakly augmented view and the student receives the strongly augmented view. Both branches extract patch features after the final LayerNorm of the ViT backbone. The CLS and register tokens are removed, and the remaining features retain their original spatial order:

$$Z^s_{bg} = F^{\text{patch}}_{\theta_s}(v^s_{bg}; M_{bg}), Z^t_{bg} = \text{sg}(F^{\text{patch}}_{\theta_t}(v^w_{bg})). \quad (6)$$

Here, $F^{\text{patch}}_{\theta_s}$ denotes the spatial-token sequence output by the backbone after its final normalization layer, $M_{bg}$ denotes the patch mask on the student side, and $sg$ denotes stop-gradient. Each sequence contains P spatial positions with d-dimensional features; for a 224 × 224 global view, $P = 256$. Because the complete geometric transformation and patch layout are identical, the $i$th teacher and student patch tokens are paired directly by index, without nearest-neighbor search, object-box matching, or additional position estimation. Masking replaces the input content at a student position but does not remove that position. Masked positions therefore retain corresponding output features and are included in the LC calculation.

The default cosine formulation introduces no additional projection head. Instead, backbone patch features on both sides are L2-normalized along the feature dimension, after which the cosine distance of each pair is computed:

$$\hat{\mathbf{z}}^s_{bgi} = \frac{\mathbf{z}^s_{bgi}}{\max(\| \mathbf{z}^s_{bgi} \|_2, \varepsilon)} \quad (7)$$

$$\hat{\mathbf{z}}^t_{bgi} = \text{sg}(\frac{\mathbf{z}^t_{bgi}}{\max(\| \mathbf{z}^t_{bgi} \|_2, \varepsilon)}) \quad (8)$$

$$d_{bgi} = 1 - (\hat{\mathbf{z}}^s_{bgi})^T \hat{\mathbf{z}}^t_{bgi} \quad (9)$$

where $\hat{\mathbf{z}}^s_{bgi}$ and $\hat{\mathbf{z}}^t_{bgi}$ denote the student and teacher features at corresponding positions, and $\varepsilon$ is a positive constant for numerical stability. The distance constrains feature direction and reduces the direct influence of feature-magnitude differences. The final LC loss is computed by taking the arithmetic mean over all samples, the two shared views, and all patch positions. It includes both visible and masked positions, does not filter for visible patches only, and does not grow through a simple summation over patches.

The teacher weak view still contains the speckle and acquisition fluctuations present in the original SAR image and must not be treated as noise-free ground truth. Teacher features are self-supervised targets produced by the EMA network rather than annotations from a clean image. Positional pairing depends only on the complete geometric transformation and fixed spatial indices; it therefore constrains the same spatial location without requiring object boxes, and not every patch should be interpreted as the same target location. LC directly enforces local feature consistency under the specified augmentation distribution. Whether this consistency further improves semantic stability and downstream robustness is evaluated through the subsequent ablation and transfer experiments.

LC and iBOT operate on the same generic backbone representation layer but differ in supervision form and spatial coverage. iBOT sends patch features through a prediction head and matches teacher target distributions only at masked positions. LC directly constrains feature directions before the projection head and covers all spatial positions in the shared views. DINO additionally provides cross-view semantic supervision through CLS tokens. The three objectives constrain cross-crop prediction, local-context inference, and same-location cross-perturbation consistency, respectively, and are optimized jointly within a unified teacher–student framework. The added design thus couples SAR-specific intensity perturbations with local feature constraints under shared geometry.

### *E. Joint Training Objective*

For an unified description of the supervision signals, let $B$ denote the number of samples in a training batch participating in each loss, $G=2$ the number of global views, $V=10$ the total number of multi-crop views, and $P$ the number of patches in one global view. During distributed training, devices apply the same reduction rule to their losses and synchronize student gradients; the neighborhood set for KoLeo is restricted to the device-local batch that actually participates in its computation. DINO takes projected CLS distributions as input, whereas iBOT takes projected patch distributions; both teacher targets are stop-gradient. KoLeo uses global CLS features from the student backbone, and LC uses backbone patch features from paired weak and strong views.

The DINO loss is a cross-view cross-entropy objective. Let $p_{bv}^{D}$ be the student CLS prediction distribution for view $v$, $q_{bg}^{D}$ the teacher target distribution for global view $g$, and $A$ the set of valid directed pairs between teacher global views and distinct student views. Two global views and eight local views yield 18 directed pairs in our setting. The loss is:

$$L_{\text{DINO}} = -\frac{1}{B|A|}\sum_{b=1}^{B}\sum_{(g,v)\in A}\sum_{c=1}^{C_D}\text{sg}(q_{bg,c}^{D})\log p_{bv,c}^{D}. \quad (10)$$

Here, $c_D$ is the output dimension of the DINO prediction head and $c$ indexes the prediction dimensions rather than predefined SAR categories. Student distributions are obtained by temperature scaling and softmax normalization of the prediction-head output. Teacher distributions follow the centering and temperature mechanism inherited from DINOv2. The set A excludes self-matches between a teacher global view and the identical student global view, while retaining matches across global views and between global and local views. Averaging over batch samples and all valid view pairs promotes consistency of prediction distributions across regions and scales.

The iBOT loss computes cross-entropy only for patch predictions at student-masked positions. Let $M_{bg}$ denote the set of masked positions in global view $g$ of sample $b$, and let $p_{bgi}^{P}$ and $q_{bgi}^{P}$ denote the student and unmasked-teacher prediction distributions at the corresponding position.

$$L_{\text{iBOT}} = -\frac{1}{BG}\sum_{b=1}^{B}\sum_{g=1}^{G}\frac{\sum_{i\in M_{bg}}\sum_{c=1}^{C_P}\text{sg}(q_{bgi,c}^{P})\log p_{bgi,c}^{P}}{\max(1,|M_{bg}|)} \quad (11)$$

where $c_P$ is the output dimension of the patch prediction head. The loss is first averaged over masked positions within each sample–view pair and then over samples and global views; a view with no masked positions contributes zero. Teacher and student patch distributions use the temperature and target-normalization mechanisms associated with their prediction heads. This reduction prevents a view with more masked positions from receiving a larger total weight solely because it contains more positions. iBOT requires the student to predict teacher distributions at missing locations from visible context, complementing the learning of local structure and contextual representations.

KoLeo regularization encourages global student features to remain dispersed on the unit hypersphere. Let $\mathbf{h}_{bg}$ be the L2-normalized global CLS feature of the student and $n(b,g)$ the nearest-neighbor index among samples other than b within the same global-view set.

$$n(b,g) = \text{argmin}_{j\neq b}\|\mathbf{h}_{bg}-\mathbf{h}_{jg}\|_2 \quad (12)$$

$$L_{\text{KoLeo}} = -\frac{1}{BG}\sum_{g=1}^{G}\sum_{b=1}^{B}\log(\|\mathbf{h}_{bg}-\mathbf{h}_{n(b,g),g}\|_2+\varepsilon) \quad (13)$$

Here, $j$ indexes candidate neighbors, $\varepsilon$ prevents numerical instability as the distance approaches zero, and each device-local batch participating in the neighbor search contains at least two samples. Neighbors are found separately within each global-view sample set, and the result is averaged over samples and views; the two views of the same sample are not placed in the same neighbor pool. KoLeo uses student CLS features before the projection head and requires no teacher target. Its role is to reduce excessive concentration in feature space. The neighbor search in Eq.12 belongs only to KoLeo inter-sample regularization and is unrelated to the fixed-index patch correspondence used by LC.

The local-consistency loss is averaged over all corresponding patches in the shared-geometry weak–strong views:

$$L_{\text{LC}} = \frac{1}{BGP}\sum_{b=1}^{B}\sum_{g=1}^{G}\sum_{i=1}^{P}[1-(\hat{\mathbf{z}}_{bgi}^{s})^{T}\text{sg}(\hat{\mathbf{z}}_{bgi}^{t})] \quad (14)$$

The normalized patch features are defined in Eq.7. Teacher features are produced by the EMA encoder from weakly augmented views, and student features are produced from strongly augmented views. The averaging range explicitly covers $B$ samples, $G$ shared global regions, and $P$ spatial positions, including masked positions. The loss does not pass through the DINO or iBOT projection heads. LC aligns feature directions at the same spatial location under the specified speckle and intensity perturbations, and mean reduction prevents the loss scale from increasing linearly with batch size, the number of shared views, or the number of patches.

The final training objective is the weighted combination of four losses:

$$L = \lambda_{\mathrm{DINO}} L_{\mathrm{DINO}} + \lambda_{\mathrm{iBOT}} L_{\mathrm{iBOT}} + \lambda_{\mathrm{KoLeo}} L_{\mathrm{KoLeo}} + \lambda_{\mathrm{LC}} L_{\mathrm{LC}} \tag{15}$$

The four nonnegative coefficients control the relative contributions of cross-view prediction, masked patch prediction, feature-distribution regularization, and local perturbation consistency. The first three objectives follow DINOv2 [9], whereas LC is introduced in this work. Their numerical weights are treated as training hyperparameters and are reported with the implementation settings. Each loss is averaged over the sample, view, and position ranges defined above; the combination of loss weight and reduction rule determines the effective contribution of each signal. Gradients are computed only for the student and its trainable projection heads. All teacher targets are stop-gradient, and the teacher is updated according to Eq.5.

### *F. Foundation-Model Pre-Training Strategy*

Pre-training follows a staged domain-adaptation strategy. We first load the optically pre-trained DINOv2 ViT-B/14 backbone, whose global semantic discrimination and local structural perception provide initialization for SAR-domain training. During the initial stage, the backbone is frozen and only the trainable projection heads are optimized, allowing their prediction distributions to adapt to SAR inputs. This stage lasts E_f epochs. The backbone is then unfrozen and jointly optimized with the projection heads so that SAR-domain supervision can update the encoder. Because the default LC objective acts directly on backbone features rather than through a projection head, it produces no trainable backbone gradients while the backbone is frozen; its representation-adaptation effect begins after unfreezing. KoLeo likewise cannot update a frozen backbone.

The backbone and projection heads use different learning rates during joint optimization. A smaller backbone learning rate limits changes to the existing visual representation, while a larger projection-head learning rate adapts the self-supervised prediction targets. Layer-wise learning-rate decay is also applied: layers closer to the input are updated with smaller steps, whereas deeper layers retain greater flexibility for domain adaptation. For a backbone with $H$ parameter groups, the learning rate of layer $l$ is as follows:

$$\eta_l(k) = \eta_{\mathrm{backbone}}(k)\, \rho^{H-l}, 0 < \rho \le 1 \tag{16}$$

where $\eta_{\mathrm{backbone}}$ is the reference learning rate of the deepest layer, $\rho$ is the layer-wise decay factor, and l increases from the input side toward the output side. The learning rate of each parameter group increases linearly during warm-up and then follows cosine decay; teacher momentum is also increased with a cosine schedule. A learning-rate schedule over one stage can be written as follows:

$$\begin{aligned} \eta(k) &= \eta_{\max} k / K_w, & 0 \le k < K_w \\ \eta(k) &= \eta_{\min} + \frac{\eta_{\max} - \eta_{\min}}{2}[1 + \cos(\pi \frac{k - K_w}{K - K_w})], & K_w \le k \le K \end{aligned} \tag{17}$$

$$m_k = 1 - \frac{1 - m_0}{2}[1 + \cos(\frac{\pi k}{K})] \tag{18}$$

where $K$ is the number of update steps in the corresponding scheduling stage, $K_w$ the number of warm-up steps, $\eta_{\max}$ and $\eta_{\min}$ the peak and terminal learning rates, respectively, and $m_0$ the initial teacher momentum. The backbone and projection heads use their own learning-rate scales, while the teacher-momentum schedule follows the self-supervised joint-training process. The freezing duration, layer-wise decay factor, warm-up length, and scheduling endpoints are specified as implementation hyperparameters in the experimental configuration.

Samples from all sources are first merged into a unified unlabeled collection, from which training batches are constructed by random sampling without additional dataset balancing. The expected sampling proportion of each source is therefore determined by the number of samples retained after cleaning and test-set isolation rather than by equal sampling probabilities across datasets. Distributed training shuffles and partitions the merged collection to exploit variations among the source SAR datasets. The pre-training collection is isolated from downstream test sets according to Section III-A; test images and labels are excluded from pre-training, hyperparameter selection, and model selection.

Pre-training uses the AdamW optimizer, with hardware and learning-rate settings summarized in Table I. After training, we export the EMA teacher backbone, retaining the patch embedding, Transformer blocks, normalization layers, and token parameters required by the generic encoder. The DINO and iBOT projection heads are pre-training modules and are not required for downstream classification or detection. Classification adaptation uses image-level representations, whereas detection adaptation uses spatial patch features with a task-specific network. Detailed downstream settings and results are presented in the next section.

TABLE I
FOUNDATION-MODEL PRE-TRAINING CONFIGURATION

| **Configuration** | **Setting** |
|---|---|
| Backbone initialization | Optically pre-trained DINOv2 ViT-B/14 with four register tokens |
| Global view and patches | 224 × 224 input; 14 × 14 patches; 256 patches |
| Number of views | Two global views and eight local views; two shared-geometry weak–strong pairs |
| Compute | 4 × NVIDIA H100 GPUs |
| Training duration and batch size | 1,000 epochs; global batch size 256; 64 samples per GPU |
| Optimizer | AdamW |
| Frozen stage | Backbone frozen for E_f epochs; projection-head learning rate 5 × $10^{-4}$ |
| Joint optimization | Initial backbone learning rate 1 × $10^{-5}$; projection-head learning rate 1 × $10^{-4}$ |
| Learning-rate and teacher schedules | Linear warm-up for 10 epochs, followed by cosine decay to 1 × $10^{-6}$; ρ = 0.9; teacher EMA momentum increased from 0.992 to 1.0 with a cosine schedule |

## IV. EXPERIMENTS

This section evaluates the representation learning capability of Xuanjian-DINOSAR on two types of downstream tasks: classification and object detection. Section IV-A details the composition of the pre-training and downstream evaluation datasets. Section IV-B describes the downstream adaptation strategies. Section IV-C reports the main experimental results. Section IV-D provides an in-depth analysis of the impacts of SAR-domain pre-training, the local semantic consistency branch under perturbations, and the downstream adaptation strategies. Section IV-E presents qualitative results and representation analysis.

### *A. Experimental Setup*

We conduct self-supervised pre-training on SAR imagery with XuanJian-DINOSAR and evaluate the transferability of the learned representations through classification and object-detection tasks.

#### *1) Pre-Training Datasets*

To increase data diversity, we aggregate multi-source SAR imagery containing ships, vehicles, aircraft, and complex scenes. The pre-training collection comprises FUSAR-Ship [20], SAMPLE [21], SAR-AIRcraft-1.0 [22], MSTAR [23], SIVED [24], FAIR-CSAR [25], RSAR [26], and the SOC-40 training split of ATRNet-STAR [27], for a total of 367,285 images used in training. These datasets complement one another in target categories, background clutter, spatial scale, target orientation, and imaging condition, providing a heterogeneous basis for generic SAR representation learning. Pre-training uses only SAR imagery and does not access class labels, target boxes, or other manual annotations. Before pre-training, all images undergo unified format conversion, invalid-sample removal, and size normalization. Table 2 summarizes the relevant information of each dataset.

TABLE II
COMPOSITION OF THE PRE-TRAINING DATA

| Dataset | Year | # Imgs. | # Targets | Size | Res. (m) | Band | Pol. | Description |
|---|---|---|---|---|---|---|---|---|
| FUSAR-Ship[20] | 2020 | 5,243 | 15 | 512×512 | 1.124×1.728 | C | DH/DV | Multiple ship classes and sea clutter |
| SAMPLE[21] | 2019 | 5,380 | 10 | 128×128 | 0.3 | X | Single | Paired simulated and measured targets |
| SAR-AIRcraft-1.0[22] | 2023 | 7400 | 7 | 800–1500 | 1 | C | Single | Fine-grained aircraft targets |
| MSTAR[23] | 1995 | 116598 | 10 | 54–193 | 0.3 | X | Single | Canonical vehicle-target data |
| SIVED[24] | 2023 | 1,044 | 1 | 512×512 | 0.1–0.3 | X/Ku/Ka | Single | Vehicle detection in complex backgrounds |
| FAIR-CSAR[25] | 2025 | 11,453 | 22 | 1024×1024 | 1–5 | C | Single/Dual | Fine-grained complex-valued SAR structure |
| SARSim | 2021 | 56234 | 12 | 512×512 | — | Multi-band | Single | Multiple target classes |
| RSAR[26] | 2025 | 95,842 | 6 | 800×800 | — | Multi-band | Multi-polarization | Multi-class oriented-object detection |
| ATRNet-STAR[27] (SOC-40 train) | 2026 | 68,091 | 40 | 128×128 | 0.12–0.15 | X/Ku | Quad | Vehicle classes and condition shifts |

#### *2) Downstream Evaluation Datasets*

The downstream evaluation covers two types of tasks: fine-grained image classification and object detection. For the classification task, ATRNet-STAR [27] is adopted. This section reports the results on the SOC-50 dataset and five Extended Operating Conditions (EOC) tasks, aiming to evaluate the model's capability in fine-grained category discrimination and cross-condition transfer. For the detection task, RSAR [26], Xuanjian-RSAR [28], and FAIR-CSAR [25] are employed to assess the applicability of the learned representations across different rotated bounding box detection benchmarks. Table 3 presents the relevant information of the evaluation datasets.

TABLE III
DOWNSTREAM TASKS AND EVALUATION DATASETS

| Dataset | Year | # Imgs. | # Targets | Size | Res. (m) | Band | Pol. | Description |
|---|---|---|---|---|---|---|---|---|
| ATRNet-STAR[27] | 2026 | 108,569 | 50 | 128×128 | 0.12–0.15 | X/Ku | Quad | Vehicle recognition and EOC evaluation |
| RSAR[26] | 2025 | 95,842 | 6 | 800×800 | — | Multi-band | Multi-polarization | General oriented-object detection |
| XuanJian-RSAR[28] | 2026 | 40605 | 26 | 1024×1024 | 1-10 | Multi-band | Multi-polarization | Fine-grained oriented-object detection |
| FAIR-CSAR[25] | 2025 | 23,825 | 22 | 1024×1024 | 1–5 | C | Single/Dual | Fine-grained complex-valued SAR detection |

ATRNet-STAR [27] is a fine-grained SAR classification dataset of ground vehicles, with categories defined at the vehicle-model level. Its SOC-40 task contains 40 civilian vehicle categories, and SOC-50 augments them with 10 ground military vehicle classes from MSTAR. The EOC tasks assess transfer across scenes, depression angles, azimuth angles, imaging bands, and polarization states. We follow the official EOC training and test splits; full protocols are given in [27].

### *B. Downstream Adaptation and Implementation Details*

We evaluate the pre-trained representations through classification adaptation and oriented-object detection adaptation. The experiments include comparisons with existing methods and comparisons among pre-training weights under the same downstream setting. The former characterize task-level performance, whereas the latter isolate the effect of the pre-training scheme. Results involving different backbones or detectors are not treated as controlled comparisons that vary only the pre-trained weights.

#### *1) Classification Adaptation*

##### *a) Linear Classification Head*

When applying the Xuanjian-DINOSAR foundation model to the classification task, the architecture adopts the foundation model as the backbone, supplemented by a linear classification layer serving as the recognition head. The linear classification layer consists of three fully connected layers: the first layer constructs a 768-dimensional input feature vector from the tokens output by the backbone, and then doubles the dimension to 1,536; the second layer performs another fully connected mapping while keeping the dimension unchanged; the third layer maps the expanded feature vector to the dimension corresponding to the number of final classification labels.

During model training, data augmentation strategies such as random horizontal flipping, random vertical flipping, random scaling, random translation, and multiplicative noise addition are introduced.

In terms of feature aggregation strategy, this study adopts the average of all patch tokens output by the backbone network as the input feature for the classification head, rather than the CLS token used in the traditional ViT architecture. Ablation experiments demonstrate that patch mean aggregation can more effectively consolidate local representations from different spatial positions, outperforming the use of the CLS token alone or hybrid weighting strategies on the current classification task.

*b) Classification Fine-Tuning Strategy*

We adopt LoRA fine-tuning as the fine-tuning scheme for the recognition task, which demonstrates high efficiency with limited data. During fine-tuning, the backbone network is completely frozen. LoRA branches are added to the weight matrices corresponding to the query projection (Q), key projection (K), and value projection (V) of the attention modules in each block. Additionally, a LoRA branch is also added at the output projection (O) after the multi-head attention aggregation. The rank of the LoRA matrix is set to 8, with $\alpha = 16$.

Label smoothing is introduced for the classification task, with the smoothing parameter set to ε = 0.15. That is, compared to one-hot encoded labels where the true class is assigned 1 and all other classes are 0, label smoothing modifies the true class label from 1 to $1-\varepsilon$, and the other class labels to $\varepsilon/(K-1)$, where $K$ is the total number of classes. This aims to improve the generalization capability of the model.

The training is conducted for 200 epochs. The learning rate for both the LoRA branches and the linear classification layer is set to $5e^{-4}$, with a cosine annealing schedule controlling the learning rate decay. To enhance training stability, the first 5 epochs are designated as warm-up epochs. Additionally, a gradient norm constraint (grad norm) is applied, limiting the maximum gradient norm to 2.0, thereby further regulating the training process and optimizing the training outcome.

*2) Detection Adaptation*

We adopt RoI Transformer as the downstream detector and conduct experiments on RSAR, XuanJian-RSAR, and FAIR-CSAR. To keep the detector architecture and training protocol consistent, we modify only the input interface and feature-output format required by the pre-trained foundation model. The detection head, feature pyramid, and remaining RoI Transformer hyperparameters are otherwise unchanged.

*a) Input-Size and Feature-Scale Adaptation*

Because the foundation model uses a fixed spatial downsampling factor, the short side of each SAR image is resized to 896 pixels so that the input resolution matches the backbone downsampling ratio. The foundation model outputs a single-scale feature map at 1/14 of the input resolution, whereas the original RoI Transformer receives multi-level features from ResNet-50. We therefore apply lightweight upsampling to the single-scale output and construct a multi-scale representation compatible with the detector interface before passing it to the original feature-pyramid network. This adaptation changes only feature resolution and interface format; it introduces no new detection branch and does not modify backbone feature extraction.

*b) Detection Network and Training Strategy*

Apart from input-size and feature-scale adaptation, the detector follows the original RoI Transformer architecture and implementation, including region proposal, rotated proposal modeling, detection heads, and loss functions.

We use two-stage progressive fine-tuning to adapt the foundation model to oriented-object detection. In the first stage, the pre-trained weights initialize the backbone, whose parameters are then frozen. Only the neck, detection heads, and upsampling layers introduced for feature-scale adaptation are trained. This stage allows the randomly initialized downstream modules to adapt to the distribution of backbone features before joint optimization, reducing instability caused by disparate parameter states at the start of training. In the second stage, the backbone is unfrozen and jointly fine-tuned end to end with the neck and detection heads, allowing the pre-trained representation to adapt to target scale, orientation, and rotated-box localization. The same two-stage procedure is used for controlled baselines under the same detector. Results involving different backbones or detection frameworks are reported only for overall task comparison and are not treated as strict ablations of the pre-trained weights.

The detector is trained for 24 epochs with an initial learning rate of $5e^{-4}$, reduced by a factor of 0.1 at epochs 12 and 18. These settings apply to the RoI Transformer adaptation used in this work and do not imply that other detection frameworks use the same schedule. The optimizer, batch size, weight decay, gradient clipping, checkpoint loading, and inference thresholds follow the public RoI Transformer implementation.

## C. Downstream Results

Classification results cover SOC-50 and five EOC tasks on ATRNet-STAR, while detection results cover RSAR, XuanJian-RSAR, and FAIR-CSAR. XuanJian-DINOSAR uses the RoI Transformer adaptation described in Section IV-B2. The comparison tables also include methods with different backbones and task frameworks to characterize downstream performance. Because such comparisons additionally vary backbone structure, feature adaptation, or training protocol, their performance differences cannot be attributed entirely to the pre-training objective. Comparisons among pre-trained weights under the same downstream setting are presented in Section IV-D.

*1) Classification Results*

Table IV reports classification accuracy on ATRNet-STAR

SOC-50. XuanJian-DINOSAR achieves 96.70%, exceeding AFRL-DINOv2 [13] by 1.80 percentage points and trailing the iTPN-B and iTPN-L variants of SARATR-X-v2 [17] by 1.10 and 1.70 points, respectively. Its SOC-50 accuracy is therefore higher than that of AFRL-DINOv2 and close to the SARATR-X-v2 results listed in the table. Because the compared methods use different backbones and pre-training schemes, the table is intended primarily as a task-level comparison.

To further evaluate transfer across imaging conditions, Table V reports five EOC tasks. EOC-Scene, EOC-Depr, EOC-Az, EOC-Band, and EOC-Pol evaluate transfer across scenes, depression angles, azimuth angles, frequency bands, and polarization states, respectively. We compare XuanJian-DINOSAR with AFRL-DINOv2 [13], SARATR-X [6], and the other methods listed in the table.

TABLE IV
COMPARISON WITH EXISTING METHODS ON ATRNET-STAR SOC-50 CLASSIFICATION ACCURACY

| Method | Year | Backbone | Accuracy (%) |
|---|---|---|---|
| ViT[29] | 2021 | ViT-B | 59.20 |
| HDANet[30] | 2023 | CNN | 63.70 |
| SARMAE[16] | 2026 | ViT-L | 81.40 |
| ConvNeXt[31] | 2022 | ConvNeXt-B | 81.60 |
| SARATR-X[6] | 2025 | HiViT-B | 85.20 |
| AFRL-DINOv2[13] | 2025 | ViT-B | 94.90 |
| SARATR-X-v2[17] | 2026 | iTPN-B | 97.80 |
| SARATR-X-v2[17] | 2026 | iTPN-L | 98.40 |
| **XuanJian-DINOSAR (Ours)** | 2026 | **ViT-B/14** | **96.70** |

TABLE V
CLASSIFICATION ACCURACY (%) ON FIVE ATRNET-STAR EOC TASKS

| Method | EOC-Scene | EOC-Depr | EOC-Az | EOC-Band | EOC-Pol |
|---|---|---|---|---|---|
| ViT | 12.90 | 30.40 | 29.00 | 65.70 | 53.60 |
| HDANet | 33.60 | 32.90 | 16.40 | 79.50 | 63.10 |
| ConvNeXt | 16.50 | 43.10 | 21.10 | 88.40 | 83.10 |
| SARATR-X | 19.50 | 39.90 | 26.40 | 89.20 | 84.60 |
| AFRL-DINOv2 | 47.60 | 63.80 | 66.00 | 97.70 | 95.50 |
| **XuanJian-DINOSAR (Ours)** | 48.30 | **71.28** | **77.20** | **98.35** | **96.56** |

XuanJian-DINOSAR obtains the highest accuracy on all five EOC tasks in Table V. Relative to AFRL-DINOv2, the improvements on EOC-Scene, EOC-Depr, EOC-Az, EOC-Band, and EOC-Pol are 0.70, 7.48, 11.20, 0.65, and 1.06 percentage points, respectively, with larger gains under cross-depression-angle and cross-azimuth evaluation. Together with the SOC-50 result, these evaluations indicate that the model retains fine-grained class discrimination while improving transfer under the evaluated condition shifts.

Table VI further reports group-wise EOC results. For the azimuth task, training uses samples from 0°–60° and testing covers the remaining angular intervals. The depression-angle task trains at 15° and tests at 30°, 45°, and 60°. The polarization task transfers from HH to HV, VH, and VV, and the scene task transfers from desert to factory, urban, and woodland backgrounds. In the azimuth task, accuracy is 82.15% over 60°–120° and 87.65% over 300°–360°, compared with 70.65%–75.78% in the remaining intervals, reflecting differences in recognition difficulty across viewing directions. Accuracy decreases from 79.93% to 59.63% as the depression angle increases from 30° to 60°. All three cross-polarization accuracies exceed 96%, whereas cross-scene accuracy ranges from 35.04% to 49.28%, with woodland being the most difficult. Under the current data and protocol, the different condition shifts therefore affect transfer unequally, and cross-scene adaptation remains the largest source of error.

*2) Detection Results*

RSAR and XuanJian-RSAR report mAP, AP50, and AP75, whereas FAIR-CSAR reports AP50. All detection benchmarks use oriented bounding-box annotations and match predictions by oriented-box intersection over union.

*a) RSAR*

As shown in Table VII, XuanJian-DINOSAR achieves 39.90 mAP, 74.30 AP50, and 37.10 AP75 on RSAR. Relative to the RoI Transformer baseline with a ResNet-50 backbone, the three metrics improve by 4.88, 7.35, and 4.45 points, corresponding to relative gains of 13.9%, 11.0%, and 13.6%.

TABLE VI
GROUP-WISE CLASSIFICATION RESULTS FOR THE EOC TASKS

| Cross-Azimuth Evaluation | | | | | |
|---|---|---|---|---|---|
| Angle | 60°-120° | 120°-180° | 180°-240° | 240°-300° | 300°-360° |
| Accuracy | 82.15% | 75.78% | 70.76% | 70.65% | 87.65% |

| Cross-Depression-Angle Evaluation | | | |
|---|---|---|---|
| Angle | 30° | 45° | 60° |
| Accuracy | 79.93% | 74.36% | 59.63% |
| **Cross-Polarization Evaluation** | | | |
| Polarization | HV | VH | VV |
| Accuracy | 96.02% | 96.44% | 97.23% |
| **Cross-Scene Evaluation** | | | |
| Scene | factory | urban | woodland |
| Accuracy | 49.28% | 48.88% | 35.04% |

TABLE VII
COMPARISON OF ORIENTED-OBJECT DETECTION RESULTS ON RSAR

| Method | Year | Backbone | mAP | AP50 | AP75 |
|---|---|---|---|---|---|
| OrientedFormer[32] | 2024 | ResNet-50 | 37.50 | 69.80 | 35.70 |
| GSDet[33] | 2025 | ResNet-50 | 35.60 | 68.00 | 33.50 |
| SARMAE[16] | 2025 | ViT-L | — | 72.20 | — |
| ViTP[34] | 2025 | ViT-L | — | 72.30 | — |
| O2-RTDETR[35] | 2026 | ResNet-18vd | 39.50 | 73.10 | 36.90 |
| SARATR-X-v2[17] | 2026 | iTPN-B | 38.70 | 74.20 | 35.70 |
| SARATR-X-v2[17] | 2026 | iTPN-L | 38.70 | 77.30 | 34.20 |
| RoI Transformer[36] (Baseline) | 2019 | ResNet-50 | 35.02 | 66.95 | 32.65 |
| **XuanJian-DINOSAR (Ours)** | **2026** | **ViT-B/14** | **39.90** | **74.30** | **37.10** |

Compared with the iTPN-B variant of SARATR-X-v2 [17], XuanJian-DINOSAR is higher by 1.20, 0.10, and 1.40 points, respectively. Relative to its iTPN-L variant, mAP and AP75 are higher by 1.20 and 2.90 points, while AP50 is lower by 3.00 points. XuanJian-DINOSAR also exceeds O2-RTDETR on all three metrics. These results show competitive overall detection performance on RSAR, with an advantage retained under the stricter IoU threshold.

The classification results fully validate the effectiveness of the conventional multi-crop self-distillation pipeline, including DINO cross-view prediction and iBOT local masked prediction, in establishing global semantic and local structural correlations. Meanwhile, the results indicate that the local consistency (LC) learning branch significantly enhances the model's robustness to variations in imaging conditions, such as pitch angle, azimuth angle, and polarization state. Through weak-to-strong view constraints under shared geometric layouts, the model successfully learns local feature representations that remain stable under speckle noise and intensity perturbations. Consequently, it achieves superior performance over existing optical pre-training and traditional SAR self-supervised methods in fine-grained classification and cross-domain transfer tasks.

*b) XuanJian-RSAR*

XuanJian-RSAR [28] is currently the largest rotated bounding box fine-grained detection dataset. We compare ten representative detectors under a unified split and evaluation protocol and report the result of XuanJian-DINOSAR adapted to RoI Transformer, providing a performance reference for this benchmark.

TABLE VIII
COMPARISON OF ORIENTED-OBJECT DETECTION RESULTS ON XUANJIAN-RSAR

| Method | AP50 | AP75 | mAP |
|---|---|---|---|
| Faster R-CNN | 44.50 | 11.50 | 18.76 |
| Rotated-FCOS | 50.60 | 19.30 | 24.13 |
| RetinaNet | 36.00 | 11.90 | 16.15 |
| R3Det | 42.90 | 10.80 | 18.15 |
| $S^2$A-Net | 45.70 | 15.00 | 20.90 |
| Oriented R-CNN | 49.40 | 17.20 | 22.70 |
| WEKAN | 52.30 | 21.10 | 25.66 |
| DenoDetv2 | 48.70 | 18.30 | 23.04 |
| ReDet | 37.40 | 12.70 | 17.26 |
| RoI Transformer (Baseline) | 51.30 | 21.00 | 25.35 |
| **XuanJian-DINOSAR (Ours)** | **54.50** | **23.10** | **28.76** |
| Improvement over baseline | +3.20 (6.2%) | +2.10 (10.0%) | +3.41 (13.5%) |

Table VIII shows that XuanJian-DINOSAR obtains 54.50 AP50, 23.10 AP75, and 28.76 mAP on XuanJian-RSAR, exceeding the other methods listed in the table on all three metrics. Compared with WEKAN, mAP increases by 3.10 points. Relative to the RoI Transformer baseline with ResNet-50, AP50, AP75, and mAP improve by 3.20, 2.10, and 3.41 points, corresponding to relative gains of 6.2%, 10.0%, and 13.5%. Together with the RSAR results, these comparisons show that the proposed model outperforms the corresponding RoI Transformer baseline on both oriented-object detection benchmarks.

*c) FAIR-CSAR*

FAIR-CSAR [25] contains fine-grained subclasses within broad categories such as aircraft, ships, bridges, and tower cranes and therefore evaluates discrimination among visually similar targets. We report AP50 for this benchmark.

TABLE IX
COMPARISON OF DETECTION RESULTS ON FAIR-CSAR (AP50)

| Method | Backbone | AP50 |
|---|---|---|
| RoI Transformer (Baseline) | ResNet-50 | 38.00 |
| SSFG-Net[37] | — | 42.60 |
| **XuanJian-DINOSAR (Ours)** | **ViT-B/14** | **42.40** |

As shown in Table IX, XuanJian-DINOSAR achieves 42.40 AP50 on FAIR-CSAR, 4.40 points above the RoI Transformer baseline with ResNet-50, corresponding to a relative gain of 11.6%. It is 0.20 points below the 42.60 reported by SSFG-Net [37]. SSFG-Net exploits joint spatial–spectral features from complex-valued single-look complex SAR imagery. With a generic pre-trained encoder and downstream detector adaptation, XuanJian-DINOSAR achieves a comparable result, indicating that its representation supports fine-grained target detection on this benchmark.

The detection results confirm the superiority of the universal encoder based on the ViT-B/14 backbone in extracting spatial features of rotated objects, particularly through the synergistic effect of the iBOT local masked prediction mechanism and the local consistency learning (LC) branch. By recovering the context of masked positions, iBOT enhances the model's perception of local scattering structures, while the LC branch effectively mitigates the interference of SAR-specific speckle noise on object localization by enforcing feature directional consistency at the same spatial position under intensity perturbations. Together, these mechanisms improve the model's fine-grained object discrimination and rotated bounding box localization accuracy in complex backgrounds, outperforming traditional CNN backbones and self-supervised models without LC constraints across multiple rotated detection benchmarks.

## D. Ablation Studies

This section aims to conduct an in-depth analysis, through ablation experiments, of the specific impacts of SAR-domain pre-training, the local consistency learning branch, and downstream adaptation strategies on downstream task performance. Table 10 first summarizes the overall comparison results for SOC-50 classification and Xuanjian-RSAR detection. For the classification task, the experiments compare the official optical pre-trained DINOv2, the SAR-domain DINOv2 model obtained by pre-training the original DINOv2 model on SAR data, and the Xuanjian-DINOSAR model incorporating the full local consistency learning branch. For the detection task, the ResNet-50 baseline based on RoI Transformer is used as the performance reference.

TABLE X
DOWNSTREAM RESULTS UNDER DIFFERENT PRE-TRAINING SETTINGS

| Method setting | SOC-50 (Accuracy, %) | XuanJian-RSAR (mAP) |
|---|---|---|
| Reference baseline (see note) | 91.76 | 25.35 |
| SAR-domain DINOv2 (without LC) | 94.01 | 26.38 |
| XuanJian-DINOSAR (Full model) | 96.70 | 28.76 |

From the overall results in Table X, it can be observed that improvements in the pre-training strategy lead to significant performance gains. On the SOC-50 classification task, compared to the official optical pre-trained weights, the SAR-domain DINOv2 model, which only introduces the conventional multi-crop self-distillation pipeline, achieves a 2.25 percentage point increase in accuracy. Building upon this, the complete Xuanjian-DINOSAR model, further incorporating the local consistency learning (LC) branch, attains an additional 2.69 percentage point improvement. This fully validates the independent contribution of the LC branch in constraining local feature consistency under speckle and intensity perturbations. On the Xuanjian-RSAR detection benchmark, the complete model achieves an mAP of 28.76, significantly surpassing the ResNet-50 baseline of 25.35. This demonstrates that the proposed pre-training scheme also possesses superior representation transfer capability in the rotated object detection scenario.

These results intuitively illustrate the complete evolutionary path from the optical baseline to SAR-domain adaptation and further to the introduction of the LC branch, fully validating the overall downstream performance of the complete pre-training scheme on both classification and detection tasks. However, to more finely deconstruct the mechanism of each module, the following sections will provide detailed analyses from the perspectives of the classification task and the recognition task, respectively.

*1) Ablation Studies on Classification*

*a) Effect of the Foundation Model on Classification*

Table XI provides a detailed comparison of the classification performance of three sets of pre-trained weights on the SOC-50 task, all of which adopt the same classification head and downstream fine-tuning strategy. The official optical pre-trained DINOv2 achieves an accuracy of 91.76%. After performing SAR-domain pre-training on the original DINOv2 model, the accuracy improves to 94.01%, indicating that SAR-domain pre-training helps enhance the applicability of optical pre-trained representations to this task. The complete model, further incorporating the local consistency learning (LC) branch, achieves an accuracy of 96.70%. This result demonstrates the independent contribution of the LC branch in improving representation quality and validates its effectiveness in enforcing local feature consistency under speckle and intensity perturbations.

TABLE XI
CLASSIFICATION RESULTS ON SOC-50 WITH DIFFERENT PRE-TRAINED WEIGHTS

| Foundation model | Official optically pre-trained DINOv2 | SAR-domain DINOv2 (without LC) | XuanJian-DINOSAR (Full model) |
|---|---|---|---|
| Classification accuracy | 91.76% | 94.01% | 96.70% |

*b) Analysis of Classification Fine-Tuning Strategies*

We further compare LoRA configurations, classifier input features, and training strategies on SOC-50. Within each comparison group, the remaining settings are held fixed to isolate the factor under study. Because the base configurations are not identical across different groups, each table should be interpreted internally rather than as a sequence of cumulative gains.

Table XII compares LoRA insertion positions. Under the current configuration, adapting Q, K, V, and O jointly yields 92.81% accuracy, compared with 92.26% for Q and V and 92.44% for Q, K, and V. Adding LoRA to the fully connected layers yields 92.62%. Table XIII compares three LoRA parameter settings: (rank = 4, α = 4), (rank = 8, α = 8), and (rank = 8, α = 16), which achieve 91.20%, 92.20%, and 92.40%, respectively. Among the tested combinations, we select insertion at Q, K, V, and O with rank = 8 and α = 16. These differences are observations under the current experimental setup and do not determine the behavior of larger parameterizations or establish whether overfitting occurs.

TABLE XII
EFFECT OF LORA INSERTION POSITIONS ON SOC-50 CLASSIFICATION ACCURACY

| LoRA insertion | Q, V | Q, K, V | Q, K, V, O | Q, K, V, O,FC |
|---|---|---|---|---|
| Classification accuracy | 92.26% | 92.44% | 92.81% | 92.62% |

TABLE XIII
EFFECT OF LORA PARAMETERS ON SOC-50 CLASSIFICATION ACCURACY

| LoRA parameters | rank=4, α = 4 | rank=8, α = 8 | rank=8, α = 16 |
|---|---|---|---|
| Classification accuracy | 91.20% | 92.20% | 92.40% |

Table XIV compares feature constructions for the classifier input, including the CLS token, the mean of patch tokens, average fusion and concatenation of CLS and mean-pooled patch features, and learnable-weighted and attention-weighted patch aggregation. The CLS token alone yields 92.59% accuracy, whereas mean-pooled patch tokens yield 93.68%, an improvement of 1.09 points. Learnable weighted pooling reaches 93.69%, only 0.01 points above arithmetic mean pooling; attention-weighted pooling reaches 93.51%, and the two CLS-fusion variants yield 92.85% and 92.80%. Simple mean pooling of patch tokens is therefore close to the best result among the tested configurations and is used to construct the classifier input. Unlike learnable weighted aggregation, arithmetic mean pooling introduces no additional aggregation parameters.

TABLE XIV
EFFECT OF CLASSIFIER INPUT CONSTRUCTION ON SOC-50 CLASSIFICATION ACCURACY

| Input feature | CLS token | patch tokens mean | CLS patch-mean | CLS patch concat | weighted patch mean | attention patch mean |
|---|---|---|---|---|---|---|
| Classification Accuracy | 92.59% | 93.68% | 92.85% | 92.80% | 93.69% | 93.51% |

Table XV compares learning-rate warm-up durations. Without warm-up and with 5- and 10-epoch warm-up, accuracy is 95.55%, 95.76%, and 95.70%, respectively; five epochs gives the highest observed value. Table XVI evaluates gradient-norm clipping. Applying clipping increases accuracy from 96.56% to 96.74%, a gain of 0.18 points. These results inform the selected training configuration. Because the differences are small, their stability should be assessed with variation across repeated runs.

TABLE XV
EFFECT OF WARM-UP DURATION ON SOC-50 CLASSIFICATION ACCURACY

| Warm-up epochs | 0 | 5 | 10 |
|---|---|---|---|
| Classification accuracy | 95.55% | 95.76% | 95.70% |

TABLE XVI
EFFECT OF GRADIENT-NORM CLIPPING ON SOC-50 CLASSIFICATION ACCURACY

| Gradient-norm clipping | Without clipping | With clipping |
|---|---|---|
| Classification accuracy | 96.56% | 96.70% |

## *E. Qualitative Results and Representation Analysis*

To further elucidate the source of the performance gain, this section provides a qualitative analysis, presenting qualitative results on both classification and detection tasks to demonstrate the practical performance of the learned representations in downstream applications.

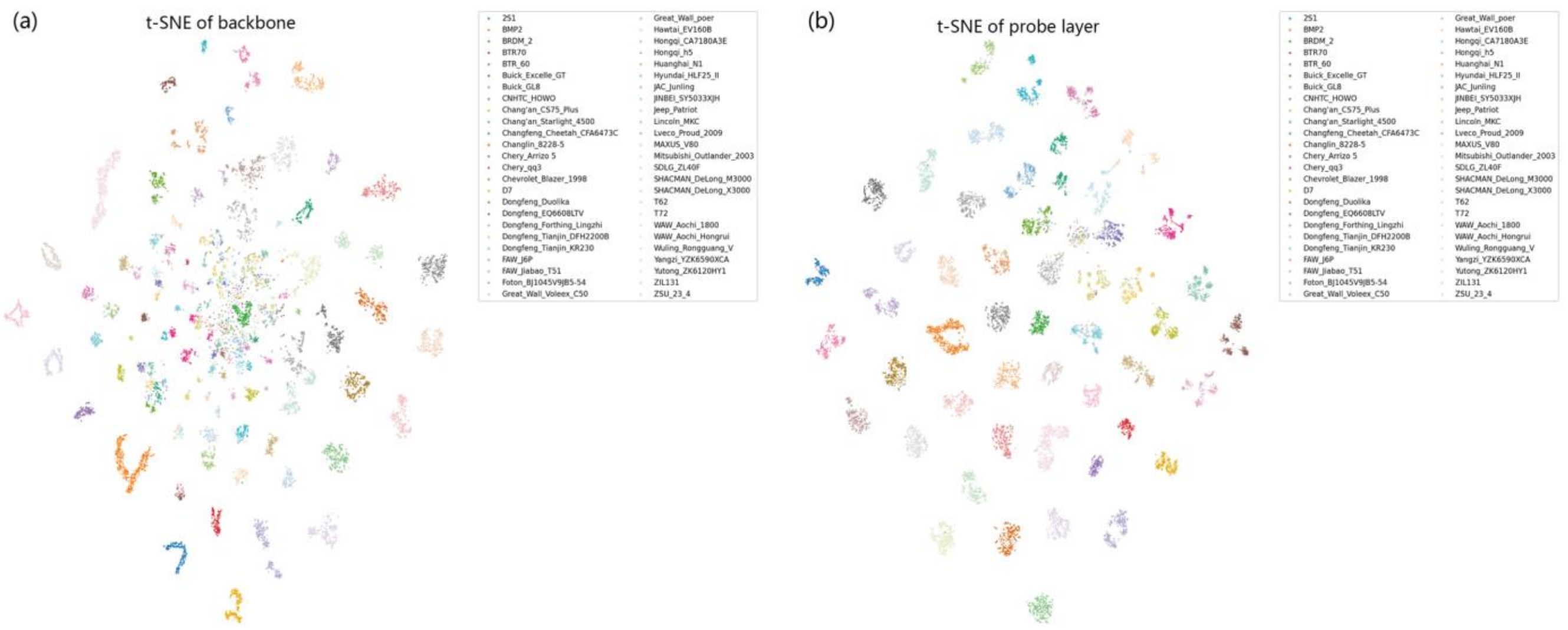

Fig. 3. T-SNE visualization of features extracted from SOC-50 class targets by the classification model. (a) shows the output feature distribution of the DINOv2 backbone network, and (b) shows the output feature distribution of the final linear hidden layer of the classification head.

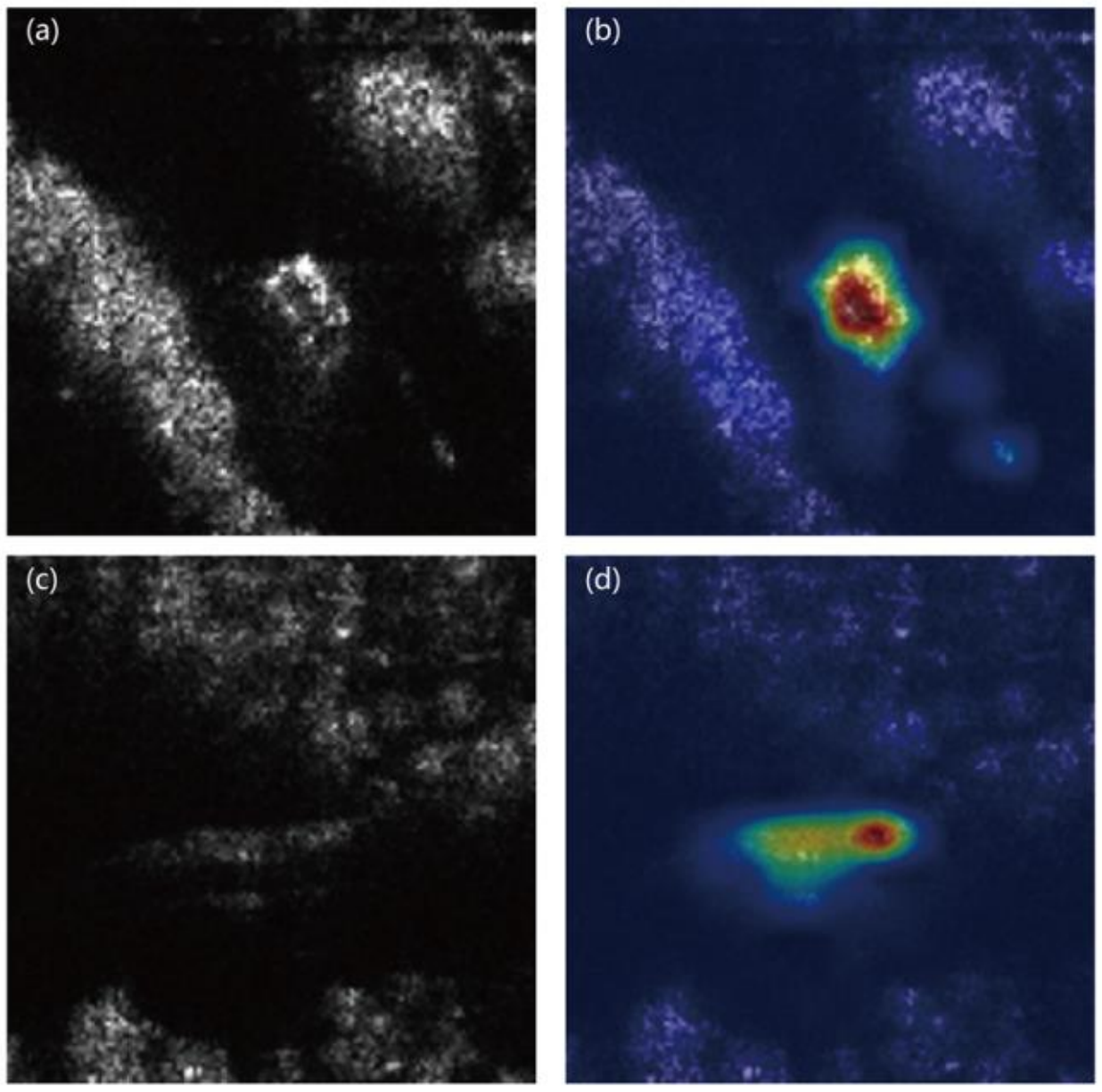

Fig. 4. Input images containing targets with background and their corresponding backbone network attention heatmaps. (a) and (c) show two example input images; (b) and (d) present the attention heatmaps of the regions focused by the backbone network corresponding to Figures (a) and (c), respectively.

*1) Visualization of Feature Distributions for Classification*

This section visualizes the output features of the Xuanjian-DINOSAR classification model to intuitively demonstrate the contribution of its feature extraction capability to the classification results. The analysis is conducted from two perspectives. First, the discriminability of the output features is examined using t-SNE visualization, which intuitively reveals the separability of high-dimensional features in the feature space. Second, the tokens output by the backbone network are mapped back to the input image in the form of attention-weighted heatmaps, allowing observation of the backbone's attention to different image regions and its ability to extract target features.

Figure 3 shows the t-SNE feature distributions of SOC-50 class targets during classification inference: (a) backbone output features (after LoRA fine-tuning) and (b) output features from the final linear hidden layer of the classification head. It can be observed that, after LoRA fine-tuning, the backbone features already exhibit strong classification capability. Most target categories are well-separated in the feature space, with only a few classes of structurally similar vehicles showing partial overlap. This result demonstrates the significant advantage of the backbone in feature extraction, indicating that the subsequent linear classification head only requires a simple design to achieve efficient fine-grained classification. As shown in Figure 3(b), the linear classification head further refines the features, ultimately achieving high separability among 50 categories and enabling fine-grained classification.

To further examine the feature extraction capability of the backbone, the output tokens are remapped to the input image according to their attention weights, producing heatmaps that indicate the backbone's attention to different regions of the input image, as shown in Figure 4. It can be seen that, for input images containing complex backgrounds, the backbone accurately focuses its attention on the target regions, effectively suppressing interference from the surrounding background. This suggests that the Xuanjian-DINOSAR foundation model concentrates more on target-inherent features during representation learning, thereby demonstrating stronger robustness and generalization ability in EOC cross-condition generalization tests.

*2) Visualization of Detection Results*

Figure 5 presents qualitative comparisons between the proposed method and mainstream baselines, including Faster R-CNN, Rotated-FCOS, and RetinaNet, under typical SAR imaging conditions. All methods adopt the same confidence threshold and post-processing pipeline. It can be clearly observed that the proposed method exhibits significant advantages in distinguishing densely arranged targets,

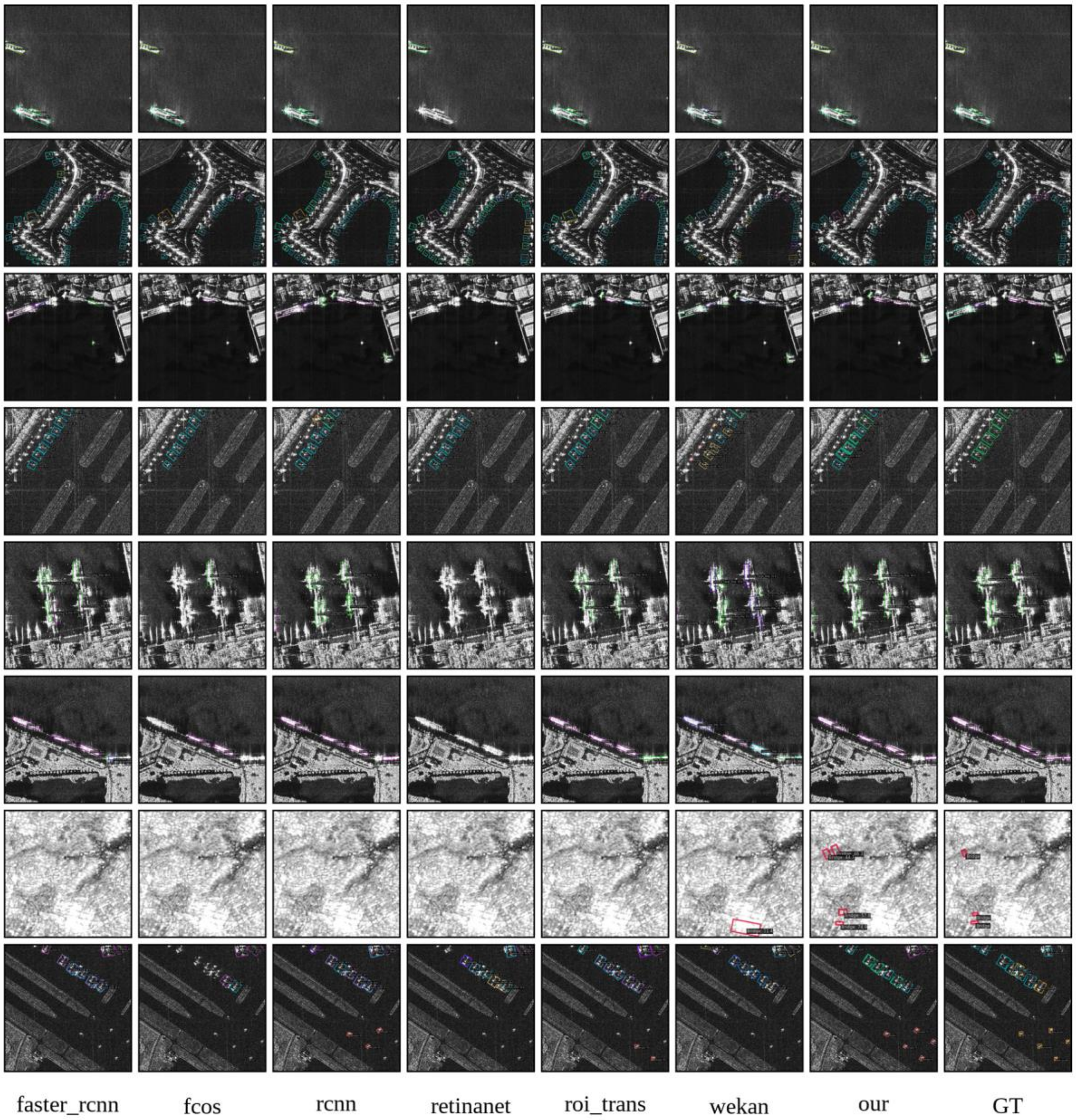

Fig. 5. Qualitative detection visualization across diverse scenes. The figure is organized as a matrix: rows represent different scene slices (e.g., scene 1, scene 2, scene 3, etc.), and columns represent different detection methods. From left to right, the columns show results from: Faster R-CNN, FCOS, R-CNN, RetinaNet, RoI Transformer, WeKAN, and our Xuanjian-DINOSAR. The rightmost column provides the corresponding ground truth annotations. Compared to other detectors, our Xuanjian-DINOSAR consistently produces more precise and robust detections, with fewer false positives and better object completeness across various scenes.

detecting small-scale objects, and suppressing background interference.

In densely berthed and occluded scenarios (e.g., rows 2, 4, and 8), baseline methods frequently suffer from missed detections due to feature confusion, or mistakenly merge adjacent targets into a single bounding box, with predicted boxes often deviating from the target center. In contrast, the proposed method accurately distinguishes ships with closely adjacent boundaries, producing detection boxes that better align with target contours without noticeable missed detections or merging. This demonstrates that the proposed method effectively enhances discriminability in dense regions during feature extraction, capturing finer inter-target boundary information, and validates the role of the local consistency learning (LC) branch in preserving local feature stability under speckle perturbations.

Furthermore, in scenarios involving small-scale targets and strong noise (e.g., rows 1 and 7), some baseline methods produce false detections under the influence of coherent speckle noise, or fail to detect small targets occupying only a few pixels. In contrast, the proposed method successfully detects these small targets while effectively suppressing false responses induced by background clutter. This result further confirms the synergistic effect of the conventional multi-crop self-distillation pipeline and the LC branch. Specifically, the iBOT local masked prediction mechanism enhances the model's perception of local scattering structures, while the LC branch enforces feature directional consistency at the same spatial position under intensity perturbations, thereby effectively mitigating the interference of SAR-specific multiplicative speckle noise on target localization. Although a small number of missed detections persist in extremely dark regions (e.g., row 3), the proposed method overall achieves significantly improved localization accuracy while maintaining a high recall rate. These qualitative results are consistent with the quantitative improvements reported in Section 4.3.

## V. CONCLUSION

This work presents XuanJian-DINOSAR, a SAR foundation model based on global and local semantic consistency, and uses it to examine how self-supervised SAR representations can balance cross-view semantic relations with local perturbation stability. Starting from an optically pre-trained DINOv2 model, self-supervised domain adaptation on multi-source SAR imagery transfers its global semantic discrimination and local structural perception to the SAR domain. We then construct shared-geometry weak – strong view pairs and impose local-consistency constraints on patch features at the same spatial locations, encouraging local representations to remain stable under intensity variation and simulated speckle perturbations. This objective complements the cross-region and cross-scale semantic consistency established by DINOv2 multi-crop learning and yields a multi-level representation-learning mechanism that spans global semantic relations and local perturbation stability.

Experiments on multiple downstream datasets covering image classification and oriented-object detection show that XuanJian-DINOSAR transfers favorably across different tasks and imaging conditions and achieves competitive performance on most reported metrics. The results suggest that combining multi-crop cross-view learning with same-location local consistency improves the adaptation of pre-trained representations to SAR imaging characteristics and downstream task variation.

More broadly, this study extends the representation objective of SAR foundation models from cross-view learning alone to coordinated global and local semantic consistency. Transferable SAR representations must establish stable high-level semantic relations across regions and scales, preserve sensitivity to local structural information, and reduce the influence of SAR-specific radiometric perturbations on feature space. XuanJian-DINOSAR provides one implementation of this principle and offers a reference for future SAR foundation models and generic representation learning from multi-source remote-sensing data.

## REFERENCES

[1] G. Gao, “Statistical modeling of SAR images: A survey,” Sensors, vol. 10, no. 1, pp. 775–795, 2010. doi: 10.3390/s100100775.
[2] Y. Hou et al., “SAR foundation models: A comprehensive review of data, models, and applications,” Science China Information Sciences, vol. 69, no. 8, Art. no. 180301, 2026. doi: 10.1007/s11432-026-5040-3.
[3] S. Chen, H. Wang, F. Xu, and Y.-Q. Jin, “Target classification using the deep convolutional networks for SAR images,” IEEE Transactions on Geoscience and Remote Sensing, vol. 54, no. 8, pp. 4806–4817, 2016. doi: 10.1109/TGRS.2016.2551720.
[4] Q. An, Z. Pan, L. Liu, and H. You, “DRBox-v2: An improved detector with rotatable boxes for target detection in SAR images,” IEEE Transactions on Geoscience and Remote Sensing, vol. 57, no. 11, pp. 8333–8349, 2019. doi: 10.1109/TGRS.2019.2920534.
[5] R. Xia et al., “CRTransSar: A visual transformer based on contextual joint representation learning for SAR ship detection,” Remote Sensing, vol. 14, no. 6, Art. no. 1488, 2022. doi: 10.3390/rs14061488.
[6] W. Li, W. Yang, Y. Hou, L. Liu, Y. Liu, and X. Li, “SARATR-X: Toward building a foundation model for SAR target recognition,” IEEE Transactions on Image Processing, vol. 34, pp. 869–884, 2025. doi: 10.1109/TIP.2025.3531988.
[7] K. He, X. Chen, S. Xie, Y. Li, P. Dollár, and R. Girshick, “Masked autoencoders are scalable vision learners,” in Proceedings of the IEEE/CVF Conference on Computer Vision and Pattern Recognition, 2022, pp. 16000–16009.
[8] M. Caron et al., “Emerging properties in self-supervised vision transformers,” in Proceedings of the IEEE/CVF International Conference on Computer Vision, 2021, pp. 9650–9660.
[9] M. Oquab et al., “DINOv2: Learning robust visual features without supervision,” Transactions on Machine Learning Research, 2024.
[10] Y. Cong et al., “SatMAE: Pre-training transformers for temporal and multi-spectral satellite imagery,” in Advances in Neural Information Processing Systems, vol. 35, 2022, pp. 197–211.
[11] X. Sun et al., “RingMo: A remote sensing foundation model with masked image modeling,” IEEE Transactions on Geoscience and Remote Sensing, vol. 61, pp. 1–22, 2023. doi: 10.1109/TGRS.2022.3194732.
[12] A. Fuller, K. Millard, and J. Green, “CROMA: Remote sensing representations with contrastive radar-optical masked autoencoders,” in Advances in Neural Information Processing Systems, vol. 36, 2023, pp. 5506–5538.
[13] N. Inkawhich, “On the status of foundation models for SAR imagery,” arXiv preprint arXiv:2509.21722v1, 2025.
[14] Y. Du, Y. Chen, L. Huang, Y. Yang, P. Ghamisi, and Q. Du, “SUMMIT: A SAR foundation model with multiple auxiliary tasks enhanced intrinsic characteristics,” International Journal of Applied Earth Observation and Geoinformation, vol. 141, Art. no. 104624, 2025. doi: 10.1016/j.jag.2025.104624.
[15] K. Wang et al., “SAMBA: A scatter-guided masked bidirectional Mamba foundation model for SAR target recognition,” arXiv preprint arXiv:2606.31668v2, 2026.
[16] D. Liu et al., “SARMAE: Masked autoencoder for SAR representation learning,” in Proceedings of the IEEE/CVF Conference on Computer Vision and Pattern Recognition, 2026, pp. 6496–6507.
[17] W. Li et al., “SARATR-X-v2: Scale-aware structural pre-training for SAR foundation models,” arXiv preprint arXiv:2607.23238v1, 2026.
[18] J. Zhou et al., “iBOT: Image BERT pre-training with online tokenizer,” in Proceedings of the International Conference on Learning Representations, 2022.
[19] W. Li et al., “Predicting gradient is better: Exploring self-supervised learning for SAR ATR with a joint-embedding predictive architecture,” ISPRS Journal of Photogrammetry and Remote Sensing, vol. 218, pp. 326–338, 2024. doi: 10.1016/j.isprsjprs.2024.09.013.
[20] X. Hou, W. Ao, Q. Song, J. Lai, H. Wang, and F. Xu, “FUSAR-Ship: Building a high-resolution SAR-AIS matchup dataset of Gaofen-3 for ship detection and recognition,” Sci. China Inf. Sci., vol. 63, pp. 1–19, 2020.
[21] B. Lewis, T. Scarnati, E. Sudkamp, J. Nehrbass, S. Rosencrantz, and E. Zelnio, “A SAR dataset for ATR development: the synthetic and measured paired labeled experiment (SAMPLE),” in Proc. SPIE Conf. Algorithms SAR Imagery, vol. 10987, 2019, pp. 39–54.
[22] Z. Wang, Y. Kang, X. Zeng, Y. Wang, T. Zhang, and X. Sun, “SARAIRcraft-1.0: High-resolution SAR aircraft detection and recognition dataset (in chinese),” J. Radars, vol. 12, no. 4, pp. 906–922, 2023.
[23] Air Force Research Laboratory, “The air force moving and stationary target recognition database,”https://www.sdms.afrl.af.mil/index.php?collection=mstar.
[24] X. Lin, B. Zhang, F. Wu, C. Wang, Y. Yang, and H. Chen, “SIVED: A SAR image dataset for vehicle detection based on rotatable bounding box,” Remote Sens., vol. 15, no. 11, p. 2825, 2023.
[25] Y. Wu, Y. Suo, Q. Meng, W. Dai, T. Miao, W. Zhao, Z. Yan, W. Diao, G.

Xie, Q. Ke, Y. Zhao, K. Fu, and X. Sun, “Fair-csar: A benchmark dataset for fine-grained object detection and recognition based on single-look complex sar images,” IEEE Transactions on Geoscience and Remote Sensing, vol. 63, pp. 1–22, 2025.

[26] X. Zhang, X. Yang, Y. Li, J. Yang, M.-M. Cheng, and X. Li, “RSAR: Restricted state angle resolver and rotated sar benchmark,” in Proceedings of the Computer Vision and Pattern Recognition Conference, 2025, pp. 7416–7426.

[27] Y. Liu, W. Li, L. Liu, J. Zhou, B. Peng, Y. Song, X. Xiong, W. Yang, T. Liu, Z. Liu, and X. Li, “ATRNet-STAR: A large dataset and benchmark toward remote sensing object recognition in the wild,” IEEE Transactions on Pattern Analysis and Machine Intelligence, vol. 48, no. 6, pp. 6735–6753, 2026.

[28] Q. Liang, T. Wu, C. Wu, et al. XuanJian-RSAR: a SAR rotated bounding box detection dataset for finegrained detection[J]. Modern Radar, 2026, 48(8): 1-5.

[29] A. Dosovitskiy et al., “An image is worth 16×16 words: Transformers for image recognition at scale,” in Proc. Int. Conf. Learn. Represent. (ICLR), 2021.

[30] W. Li, W. Yang, W. Zhang, T. Liu, Y. Liu, and L. Liu, “Hierarchical disentanglement-alignment network for robust SAR vehicle recognition,” IEEE J. Sel. Topics Appl. Earth Observ. Remote Sens., vol. 16, pp. 9661–9679, 2023.

[31] Z. Liu, H. Mao, C.-Y. Wu, C. Feichtenhofer, T. Darrell, and S. Xie, “A ConvNet for the 2020s,” in Proc. IEEE/CVF Conf. Comput. Vis. Pattern Recognit. (CVPR), 2022, pp. 11976–11986.

[32] J. Zhao, Z. Ding, Y. Zhou, H. Zhu, W.-L. Du, R. Yao, and A. El Saddik, “OrientedFormer: An end-to-end transformer-based oriented object detector in remote sensing images,” IEEE Trans. Geosci. Remote Sens., vol. 62, pp. 1–16, 2024.

[33] Z. Ding, J. Zhao, Y. Zhou, W.-L. Du, H. Zhu, and R. Yao, “GSDet: Gaussian splatting for oriented object detection,” in Proc. 34th Int. Joint Conf. Artif. Intell. (IJCAI), 2025, pp. 900–908.

[34] Y. Li, Y. Zhang, W. Tang, Y. Dai, M.-M. Cheng, X. Li, and J. Yang, “Visual instruction pretraining for domain-specific foundation models,” arXiv preprint arXiv:2509.17562, 2025.

[35] Z. Ding, Y. Zhou, J. Zhao, W.-L. Du, X. Li, R. Yao, and A. El Saddik, “Real-time oriented object detection transformer in remote sensing images,” IEEE Trans. Geosci. Remote Sens., 2026.

[36] J. Ding, N. Xue, Y. Long, G.-S. Xia, and Q. Lu, “Learning RoI Transformer for oriented object detection in aerial images,” in Proc. IEEE/CVF Conf. Comput. Vis. Pattern Recognit. (CVPR), 2019, pp. 2849–2858.

[37] Y. Suo, Y. Wu, W. Diao, Q. Meng, X. Sun, and K. Fu, “Spatial–spectral feature guidance network for fine-grained object detection and recognition on complex-valued SAR images,” ISPRS J. Photogramm. Remote Sens., vol. 236, pp. 81–98, 2026, doi: 10.1016/j.isprsjprs.2026.03.047.